\documentclass[times,twocolumn,final]{elsarticle}

\usepackage{medima}
\usepackage{framed,multirow}

\usepackage{amssymb}
\usepackage{latexsym}

\usepackage{url}
\usepackage{xcolor}

\usepackage{hyperref}

\usepackage{subcaption}
\usepackage{bm}
\usepackage{multirow}
\usepackage{algorithm2e}
\usepackage{amsmath}
\usepackage{booktabs} 
\usepackage{lineno}
\usepackage{comment}

\usepackage[normalem]{ulem} 

\newcommand{\reviadd}[1]{\textcolor{red}{\protect#1}}
\newcommand{\reviiadd}[1]{\textcolor{blue}{\protect#1}}
\newcommand{\reviiiadd}[1]{\textcolor{purple}{\protect#1}}
\newcommand{\revidel}[1]{\textcolor{red}{\sout{\protect#1}}}
\newcommand{\reviidel}[1]{\textcolor{blue}{\sout{\protect#1}}}
\newcommand{\reviiidel}[1]{\textcolor{purple}{\sout{\protect#1}}}

\renewcommand{\reviadd}[1]{#1}
\renewcommand{\reviiadd}[1]{#1}
\renewcommand{\reviiiadd}[1]{#1}
\renewcommand{\revidel}[1]{}
\renewcommand{\reviidel}[1]{}
\renewcommand{\reviiidel}[1]{}

\renewcommand\sout[1]{}

\journal{arXiv}

\begin{document}

\verso{Jingru Fu \textit{et~al.}}

\begin{frontmatter}

\title{Synthesizing Individualized Aging Brains in Health and Disease with Generative Models and Parallel Transport}%

\author[1]{\href{https://orcid.org/0000-0003-4175-395X}{\includegraphics[scale=0.06]{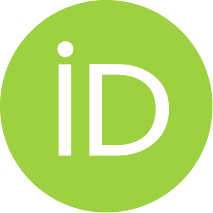}\hspace{1mm}}Jingru \snm{Fu}\corref{cor1}}
\cortext[cor1]{Corresponding author}
\ead{jingruf@kth.se}
\author[1,4]{\href{https://orcid.org/0009-0003-4183-0633}{\includegraphics[scale=0.06]{orcid.pdf}\hspace{1mm}}Yuqi \snm{Zheng}}
\ead{yuqizh@kth.se}
\author[5]{\href{https://orcid.org/0000-0003-1427-6406}{\includegraphics[scale=0.06]{orcid.pdf}\hspace{1mm}}Neel \snm{Dey}}
\ead{dey@csail.mit.edu}
\author[2,3]{\href{https://orcid.org/0000-0001-9522-4338}{\includegraphics[scale=0.06]{orcid.pdf}\hspace{1mm}}Daniel \snm{Ferreira}}
\ead{daniel.ferreira.padilla@ki.se}
\author[1,6]{\href{https://orcid.org/0000-0001-5765-2964}{\includegraphics[scale=0.06]{orcid.pdf}\hspace{1mm}}Rodrigo \snm{Moreno}\corref{cor1}}
\ead{rodmore@kth.se}

\address[1]{Division of Biomedical Imaging, KTH Royal Institute of Technology, Stockholm, Sweden}
\address[2]{Division of Clinical Geriatrics, Center for Alzheimer Research, Karolinska Institute, Stockholm, Sweden} 
\address[3]{Facultad de Ciencias de la Salud, Universidad Fernando Pessoa Canarias, Las Palmas, Spain}
\address[4]{Division of Gene Technology, KTH Royal Institute of Technology, Stockholm, Sweden}
\address[5]{Computer Science and Artificial Intelligence Laboratory, Massachusetts Institute of Technology, MA, USA}
\address[6]{MedTechLabs, BioClinicum, Karolinska University Hospital, Solna, Sweden}


\begin{abstract}
Simulating prospective magnetic resonance imaging (MRI) scans from a given individual brain image is challenging, as it requires accounting for canonical changes in aging and/or disease progression while also considering the individual brain's current status and unique characteristics. While current deep generative models can produce high-resolution anatomically accurate templates for population-wide studies, \reviiadd{their ability to predict future aging trajectories for individuals remains limited, particularly in capturing subject-specific neuroanatomical variations over time.}
In this study, we introduce Individualized Brain Synthesis (\textbf{InBrainSyn}), a framework for synthesizing high-resolution \textit{subject-specific} longitudinal MRI scans that simulate neurodegeneration in both Alzheimer's disease (AD) and normal aging. InBrainSyn uses a parallel transport algorithm to adapt the population-level aging trajectories learned by a generative deep template network, enabling individualized aging synthesis. As InBrainSyn uses diffeomorphic transformations to simulate aging, the synthesized images are topologically consistent with the original anatomy by design.
We evaluated InBrainSyn both quantitatively and qualitatively on AD and healthy control cohorts from the Open Access Series of Imaging Studies - version 3 dataset. Experimentally, InBrainSyn can also model neuroanatomical transitions between normal aging and AD. An evaluation of an external set supports its generalizability. Overall, with only a single baseline scan, InBrainSyn synthesizes realistic 3D spatiotemporal T1w MRI scans, producing personalized longitudinal aging trajectories. The code for InBrainSyn is available at \url{https://github.com/Fjr9516/InBrainSyn}.
\end{abstract}


\end{frontmatter}


\section{Introduction}
The ability to predict future individual brain trajectories holds great promise for providing valuable insights to clinicians and researchers by providing the estimation of local volume changes in brains. For example, such capabilities can formulate hypotheses on the temporal dynamics of aging and disease \citep{ziegler2012models, khanal2016biophysical}. However, achieving accurate predictions at the individual level is challenging, given substantial inter-subject variation \citep{xia2021learning}. Deep generative models have recently proven effective in generating high-quality population-level spatiotemporal atlases (also known as templates) for studying longitudinal advancements in large biomedical imaging datasets \citep{dalca2019learning, Dey_2021_ICCV}. However, existing population-level solutions offer only on-average trajectories of disease or aging for a given population and are limited in their predictive power for a specific individual subject. To that end, given only a single brain scan, we propose Individualized Brain Synthesis (\textbf{InBrainSyn}), a framework that predicts individualized spatiotemporal brain aging trajectories in both normal aging and disease. To do so, we develop a method that adapts a cross-sectional population-level spatiotemporal atlas model to individual brain scans using parallel transport.

\noindent\textbf{Motivation.} Although scalar values derived from MRI scans (e.g., cortical thickness and volume) can guide treatment in critical neurodegenerative disorders, imaging itself still serves the unique role of understanding properties that are not represented in scalar values. For example, shape changes in brain structures often precede detectable neurodegeneration, with overt brain volume reductions only manifesting in later disease stages \citep{cury2016spatio, Cury2019}. While large Alzheimer's Disease (AD) datasets have emerged to facilitate our understanding of disease progression, collecting longitudinal data presents challenges due to logistics, expense, and the common occurrence of missing data \citep{thung2016identification, pathan2018predictive, fu2023fast}. In the context of AD, simulating individualized brain shape changes based on healthy status is important because both AD and normal aging are associated with varied levels of brain atrophy~\citep{fjell2009one, whitwell2008rates, raz2005regional, fu2023deformation}. Being able to predict differences according to different health statuses can help neuroradiologists test hypotheses regarding localized neuronal degradation and resultant brain shape changes. It also could serve as “virtual control” when drug treatment is applied.

\noindent\textbf{Generative models for longitudinal synthesis.} Several studies on AD and aging employ deformable image registration to quantify geometric changes through pairwise warping between MRI images. These methods establish one or more spatiotemporal brain templates to predict the on-average trajectories of a population, studying the overall trends of the disease \citep{huizinga2018spatio, dalca2019learning, Dey_2021_ICCV}. However, AD and aging are both highly heterogeneous processes with wide inter-individual variability \citep{bagarinao2022reserve, walhovd2005effects, eavani2018heterogeneity, wrigglesworth2023health, ferreira2017cognitive}, making clear a strong need for individual-level subject-specific aging models. In recent years, deep generative models have been used to simulate and predict human brains at retrospective and prospective time points using existing (e.g. baseline) scans \citep{ravi2019degenerative, ravi2022degenerative, xia2021learning, fu2023fast, rakic2020anatomical, pombo2023equitable}. For instance, \citet{ravi2019degenerative} and \citet{ravi2022degenerative} introduced models to simulate 2D and 3D brains, respectively, employing generative adversarial training and a range of spatiotemporal and biologically informed constraints to ensure the realism of the generated images. However, their models require fine-tuning on each new scan to obtain individualization after the overall model training, which is infeasible in practice. Moreover, these methods require longitudinal data, the lack of which is a known issue in brain MRI datasets. \citet{xia2021learning} proposed a brain 2D MRI simulator that does not require longitudinal data. They propose an autoencoder framework and embed health status and age information in its latent space. However, these adversarial formulation-based models do not impose explicit anatomical constraints, which can result in unrealistic structural changes that deviate from known biological patterns. The lack of control complicates the assessment of their reliability, limiting their practical utility in longitudinal studies.

\begin{figure}[t]
    \centering
    \includegraphics[width=0.48\textwidth]{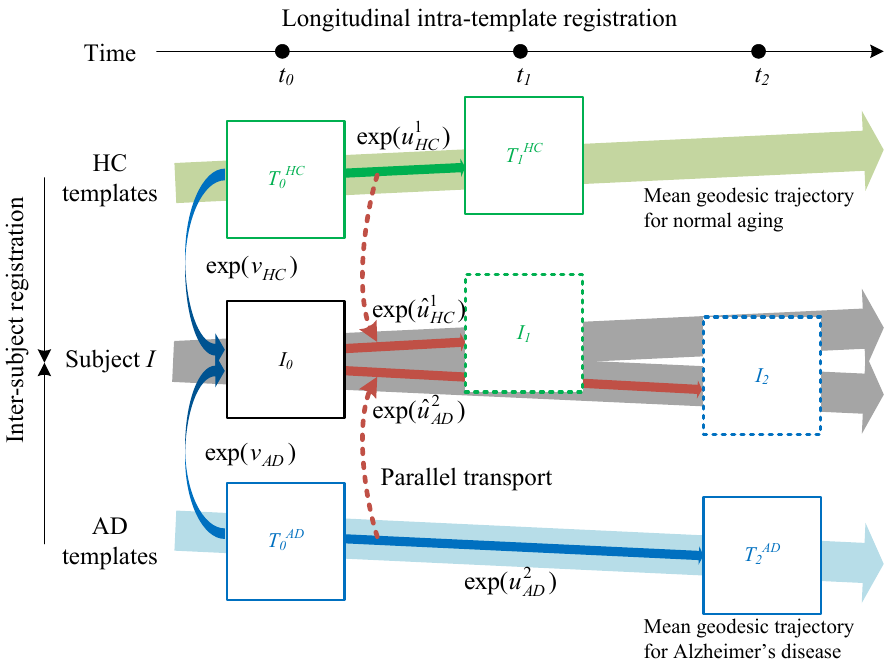}
    \caption{Schematic representation of the InBrainSyn framework, demonstrating the simulation of normal aging and Alzheimer's disease progression from a single subject observation. It comprises of two main steps: 1) learning cohort-level templates $T$ from healthy control (HC) or AD subjects, and 2) transporting morphological changes from the cohort level to the subject level using parallel transport. Canonical patterns from the given subject's age $t_0$ to age $t_i$ (i.e., $i$ can be 1 or 2 in the example, $t_2>t_1$) for normal aging and AD trajectories are captured through longitudinal intra-template registration and parametrized as the exponential of the stationary velocity field (SVF) $\bm u$ (i.e., $\text{exp}(\bm u_{HC}^{i})$ or $\text{exp}(\bm u_{AD}^{i})$). Subsequently, these patterns are parallel transported to the target subject $I_0$ along the curve from the age-matched templates $T_0^{HC}$ and $T_0^{AD}$ to the subject, which is parameterized using SVF $\bm v$ obtained through inter-subject registration (i.e., $\text{exp}(\bm v_{HC})$ or $\text{exp}(\bm v_{AD})$). The transported $\bm u$ is indicated by $\hat{\bm u}=\Pi_{BCH}(\bm u,\bm v)$, where BCH stands for Baker-Campbell-\reviadd{Hausdorff} formula. The picture shows the transport of the normal aging trajectory from time $t_0$ to $t_1$ to simulate an individual scan at $t_1$, followed by the transport of the AD trajectory from time $t_0$ to $t_2$ to simulate an individual scan at $t_2$. The solid boxes indicate the given templates or subject scan, the dashed boxes indicate the simulated ones.}
    \label{fig:InBrainSyn_with_PT}
\end{figure}

\noindent\textbf{Diffeomorphic registration for longitudinal synthesis.} A diffeomorphism is a smooth, bijective, and invertible function, widely used for medical image analysis and registration. In our context, diffeomorphic transformations ensure that generated images reflect normal anatomical changes (e.g., tissue boundaries remain continuous, and distinct anatomical regions do not merge or fragment). Previous work by \citet{fu2023fast} showed the potential of diffeomorphic registration for brain aging simulation. They used diffeomorphic registration to interpolate between two MRI scans with the maximum age gap from the same individual in the dataset. However, this setup cannot handle more flexible situations, such as having only one available scan from a subject during the inference stage. The idea of using diffeomorphic registration for brain MRI simulations is also shown in \citet{rakic2020anatomical}, where they simulate aging based on input attributes and constrain the learning using diffeomorphic registration. However, the success of this model relies on obtaining longitudinal data for training, and thus, true changes at the individual level are counted. A recent work \citet{pombo2023equitable} also used a similar strategy to infer counterfactual images under different conditions. It introduced diffeomorphic deformations in the format of starGAN \citep{choi2018stargan}, a model capable of multi-domain image-to-image translation, to ensure the anatomical plausibility of synthesis. Although their work was developed and evaluated in a down-sampled space, they showed the usefulness of using diffeomorphisms in simulations for medical images.

\noindent\textbf{Contributions.} We propose InBrainSyn, a framework that synthesizes potential brain aging trajectories in MRI in both normal aging and disease. InBrainSyn generates personalized predictions based on a single given image while also explicitly ensuring the anatomical plausibility of generated images by using diffeomorphic deformations. A schematic representation of the proposed framework, depicting the simulation of future normal aging or AD progression based on an observed subject, is presented in Fig.~\ref{fig:InBrainSyn_with_PT}. Our approach develops a deep cohort-level generative model to learn canonical normal aging and AD progression trajectories using only cross-sectional data. These population-level trajectories are then used for individualized aging prediction using an efficient parallel transport algorithm. Our contributions enable the integration of population models to subject-specific brain synthesis while remaining computationally efficient by working on the deformation field. 
As we transport deformations from population models to individuals with diffeomorphic constraints on the deformations, our predicted trajectories are both anatomically plausible and are free of the intensity artifacts that commonly occur in deep generative models.

\section{Related Work}
Researchers often leverage longitudinal neuroimaging datasets representing heterogeneous populations to investigate neurodegenerative diseases. For example, datasets like the Alzheimer's Disease Neuroimaging Initiative (ADNI) \citep{petersen2010alzheimer} and the Open Access Series of Imaging Studies (OASIS) \citep{lamontagne2019oasis}, dedicated to AD research, have tracked populations for decades. Through these expansive datasets, researchers can compare and analyze differences across populations or sub-populations, i.e., by establishing study-specific templates \citep{AVANTS2004S139, joshi2004unbiased, avants2010optimal, dalca2019learning, Dey_2021_ICCV}.

The primary focus of template establishment has revolved around the development of image mapping algorithms \citep{miller2001group, thompson2002framework, rohlfing2003extraction} capable of estimating the optimal spatial transformation to align the structures of interest, i.e., between a fixed and moving image pair. Template creation for large-scale neuroimaging data poses challenges due to the high dimensionality and volume of general neuroimaging images. Automatic image mapping or registration algorithms have streamlined the creation of brain atlases/templates, reducing the need for extensive manual intervention, such as landmark annotation \citep{lorenzen2006multi, allassonniere2007towards, wu2011sharpmean}. While traditional automatic image registration methods iteratively solve optimization problems, such as LDDMM \citep{beg2005computing} and SyN \citep{avants2008symmetric}, they can be computationally expensive and slow in practice. Recent advances in unsupervised registration networks have led to learning-based methods for generating deformable templates \citep{de2017end, 8633930, kim2021cyclemorph, chen2022transmorph}. For example, \citet{dalca2019learning} introduced conditional templates using a generator linked with a DL-based registration network to learn the deformable template of each (sub-)population. More recent work, such as \citet{Dey_2021_ICCV}, incorporates generative adversarial learning and FiLM~\citep{perez2018film} conditioning to estimate sharp and central spatiotemporal templates across populations. While robust template creation methods exhibit the potential to learn valuable representations of spatiotemporal changes from a population, when the focus shifts to subject-specific aging, preserving individualization remains unsolved. Despite the shared anatomy of the human brain, cortical folding patterns (e.g., gyri and sulci) are unique to an individual \citep{demirci2023consistency} and require distinct modeling for estimating aging trajectories. 

Existing brain MRI simulators for individuals can be categorized into two fundamental types: biophysical/biomechanics-based and data-driven/learning-based models. Biophysical/biomechanics-based methods derive brain displacement fields from prescribed atrophy values estimated by, for example, segmentation or registration. Atrophy values are incorporated into a strain energy function that factors in material properties, loading conditions, and boundary constraints \citep{khanal2016biophysical, silva2021distinguishing}. This type of method mainly uses biomechanical assumptions to reason about changes at the image level, often necessitating the redesign of assumptions based on specific problems, e.g., using different material assumptions for diseased tissues. It does not directly learn spatiotemporal neuroanatomical changes given observations. Moreover, this type of method also has a high computational cost, such as in \citep{khanal2017simulating}, 
and there is a trade-off between image resolution, feasible computation time, and computational resources. Data-driven/learning-based methods leverage deep learning techniques to directly extract insights from data to model organ evolution or disease progression. Among learned frameworks, Generative Adversarial Networks (GANs) \citep{goodfellow2014generative} have shown strong performance for generative neuroimaging.

Several GAN-based methods have been introduced for modeling aging and the progression of AD \citep{wegmayrGenerativeAgingBrain2019a, bowles2018modelling, kim2021longitudinal, xia2021learning, ravi2022degenerative, pombo2023equitable}. Training GANs using 3D brain images is normally challenging due to the high dimensionality of these images, training instability, and the sparse temporal distribution of current longitudinal brain datasets \citep{ravi2022degenerative}. To mitigate these issues, some approaches simplify the problem significantly by considering only a single 2D slice per subject \citep{wegmayrGenerativeAgingBrain2019a, xia2021learning} or downsampling the original images during simulation \citep{ravi2022degenerative}. However, these strategies may lead to a suboptimal understanding of inherently 3D anatomy and not scale to true 3D neuroimaging tasks. To address these limitations, \citet{jung2021conditional} introduced a technique to synthesize high-quality 3D medical images by incorporating a 3D discriminator into a conventional 2D GAN architecture. Nevertheless, keeping the internal consistency of 3D MRI scans using this adversarial way poses a challenge since it is difficult to assess explicitly and potentially compromises their anatomical plausibility. More recently, \citet{ravi2022degenerative} introduced a 3D training consistency mechanism and a 3D super-resolution module, representing the state-of-the-art in simulating subject-specific aging and disease progression. To address individualization, they introduced a transfer learning strategy after training, although this does not enable the direct prediction of new data without fine-tuning. Another strategy for ensuring individualization is using a reconstruction loss, as shown by \citet{xia2021learning}, which proposed a brain 2D MRI simulator that does not require longitudinal data. It utilizes an encoder-decoder framework, embedding health status and age information in the latent space. 

\reviadd{More recently, diffusion models have shown superior performance over GAN-based models in generating high-quality images, yet few studies have adapted diffusion models for this task \citep{yoon2023sadm,puglisi2024enhancing,puglisi2025}. To efficiently handle the high-dimensional 3D nature of the problem, a more straightforward approach is to use a latent diffusion model, as demonstrated in \citet{puglisi2024enhancing}. In this framework, the final image is reconstructed separately after predicting a robust latent space representation.}

In addition to high computational cost and individualization challenges, another challenge faced by deep generative approaches is maintaining biomedical plausibility. For example, \citet{ravi2022degenerative} proposed a biologically-informed constraint, aligning MRI image intensity with a monotonically decreasing pattern \citep{vemuri2010serial}. The authors attempted to resolve this problem further by using generative adversarial learning with the guidance of real distribution. However, anatomical plausibility is not explicitly constrained in adversarial learning in that neuroanatomical topology is not necessarily preserved, an important aspect for downstream tasks like morphometric studies \citep{lorenzi2015disentangling, hadj2016longitudinal, sivera2020voxel}. \citet{fu2023fast} introduced a straightforward method using deep diffeomorphic registration to interpolate temporally intermediate MRI images between two given scans, explicitly enforcing anatomical plausibility. However, this setting is inapplicable to scenarios where only a single scan is available. Recent work \citep{pombo2023equitable} employed a similar strategy, integrating diffeomorphic deformations akin to starGAN \citep{choi2018stargan}, a model capable of multi-domain transfer with domain labels to ensure realism. StarGAN model normally requires training on a sufficiently large dataset to ensure adequate representations in each subdomain, posing practical challenges since large training data is normally unavailable for many cases. With the aforementioned concerns, we found it necessary to develop a more flexible individualized brain synthesis model.

Our work is inspired by traditional deformation-based morphometric methods that statistically identify and characterize structural differences across populations or establish correlations between brain shapes using deformable registration \citep{ashburner2000voxel}. After obtaining a study-specific template, parallel transport is normally employed to extract intra-subject changes and reduce inter-subject variation into a unified template space, facilitating subsequent group comparisons. Consequently, different groups can be compared and analyzed consistently, irrespective of their initial poses \citep{younes2008transport}. However, unlike conventional subject-to-template parallel transport in morphometric methods, we use parallel transport in a reverse way in that we aim to apply the observed canonical template space changes to a given individual. Notably, parallel transport has found extensive utility in the morphological investigation of neurodegenerative conditions (most notably AD \citep{lorenzi2015disentangling, hadj2016longitudinal, sivera2020voxel}), but, to our knowledge, it has not yet been used for individualized aging synthesis in a fast DL setting.

\begin{figure*}[!t]
    \centering
    \includegraphics[width=\textwidth]{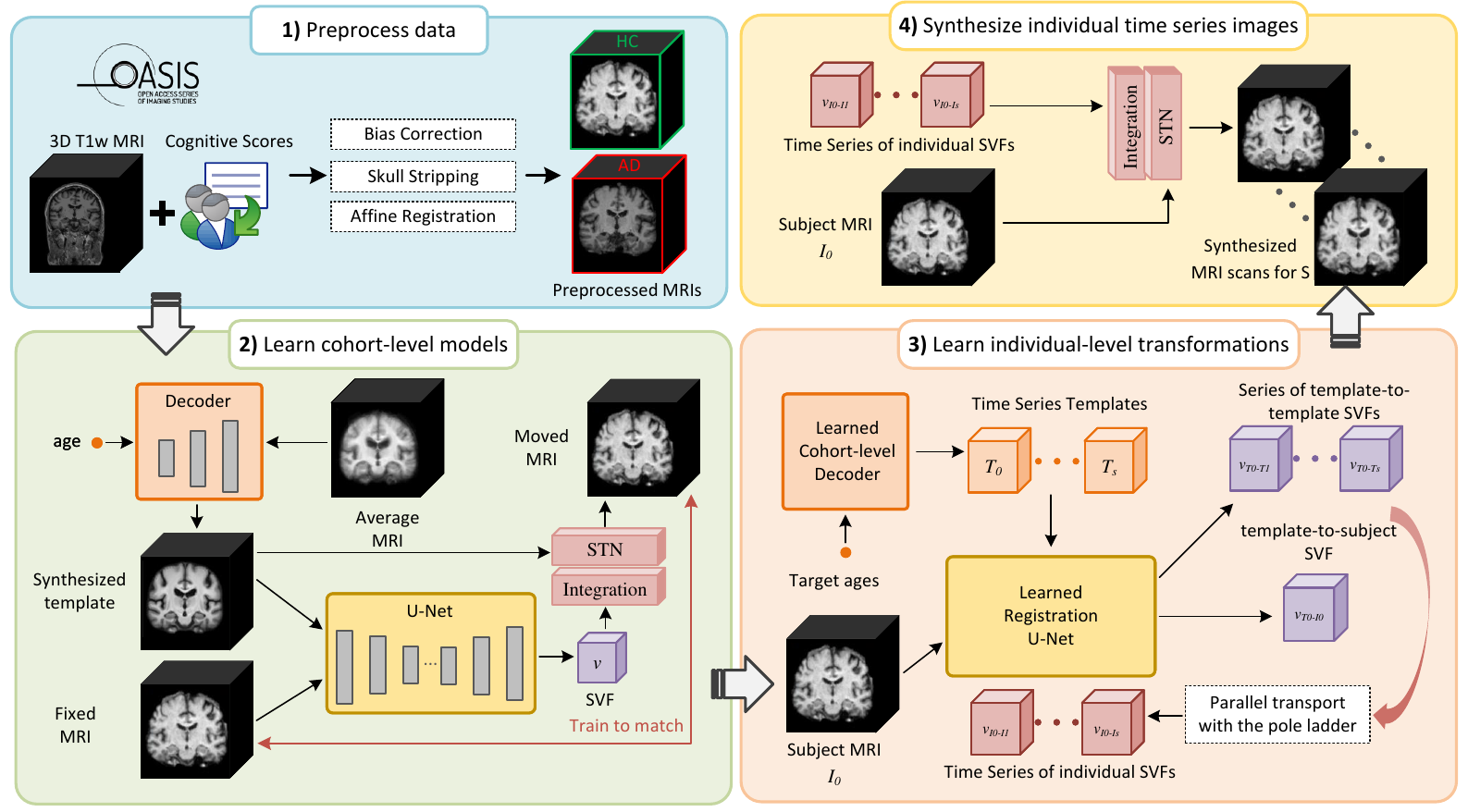}
    \caption{\textbf{InBrainSyn}. (1) \textbf{Preprocessing}: 3D T1w MRI scans and their corresponding clinical scores were collected and processed following standard protocols and partitioned into healthy control (HC) and AD groups; (2) \textbf{Learn cohort-level models}: A cohort-level decoder and a registration U-Net are trained using AtlasGAN for HC and AD cohorts to learn spatiotemporal cohort-specific neurodegeneration models, respectively; (3) \textbf{Estimate subject-specific transformations}: given a single scan $I_0$ at age $t_0$ from subject $S$, we first generate an age-matched template $T_0$ and a series of templates from other target ages (e.g., from $t_1$ to $t_s$) using the learned cohort-level decoder. We then obtain $s$ cohort-level deformations (i.e., SVFs) and a template-to-subject SVF using the learned registration U-net. Finally, we parallel transport those cohort-level deformations along the template-to-subject SVF, and the individual SVFs are obtained; (4) \textbf{Synthesize subject-specific time series images}: The final synthesized individual scans are obtained by integrating the estimating SVFs and using the resulting displacements to warp a given MRI scan.}
    \label{fig:pipeline}
\end{figure*}

Concluding our review of past work, we summarize three common challenges one must face to develop a successful brain MRI simulator: i) High computational cost; ii) Preservation of individualization; and iii) Maintenance of anatomical plausibility. In response to these challenges, we propose the InBrainSyn that effectively addresses these three concerns as detailed in the sections below. Our approach first utilizes a deep cohort-level template generation model to obtain canonical aging and AD progression trajectories. These trajectories are then transferred to individuals using parallel transport. Instead of directly working on high-dimensional scans, we generate images using a downsampled deformation field derived from a DL-based diffeomorphic registration model that maintains topological consistency.

\section{Methods}
\subsection{Framework of InBrainSyn}

Fig.~\ref{fig:pipeline} illustrates a preprocessing phase followed by the three main steps of InBrainSyn. Initially, raw T1w MRI scans and clinical scores (such as Clinical Dementia Rating (CDR) scores) were gathered and assigned to each subject \citep{lamontagne2019oasis}. Subsequently, standard preprocessing procedures, as commonly delineated in previous neuroimaging studies \citep{8633930, 9552865, Dey_2021_ICCV} were applied to the scans, including bias field correction, skull stripping, and affine standardization through FreeSurfer \footnote{\url{https://surfer.nmr.mgh.harvard.edu/}}.
In the second phase, we use the diffeomorphic registration-based framework of \citet{Dey_2021_ICCV} (here referred to as AtlasGAN), which is a state-of-the-art deep deformable spatiotemporal template generation model to create population-level spatiotemporal deformable templates that exhibit realistic anatomical features for two target cohorts: the healthy control (HC) cohort and the AD cohort. 
AtlasGAN incorporates a registration sub-network, ensuring precise alignment and the generation of diffeomorphic deformations by following the stationary velocity field (SVF) framework. Template synthesis is performed by a decoder that takes age and/or disease conditions as input.
In the third phase, we propose a method to derive individual-level transformations from the cohort-level cross-sectional deformations and the template-to-subject correspondence encoded within the template-to-subject SVF. 
To obtain these SVFs, the learned registration U-Net within AtlasGAN is used.
In the last phase, once the individual-level SVFs are obtained, they can be integrated and applied to individualize the cohort-level morphological changes over time. Further details regarding the three main phases are explained in subsequent sections.

\subsubsection{Learn Cohort-Level Template Creation and Registration Models}\label{sec:atlasgan}
Accurately capturing population-level voxel-wise morphological changes associated with normal brain aging and AD progression requires consideration of both spatial (inter-subject variation) and temporal (intra-subject variation) dynamics. In the context of \textit{deformable templates}, the ideal template is defined as an unbiased barycentric representation of the target (sub-)population \citep{AVANTS2004S139, joshi2004unbiased, avants2010optimal, Dey_2021_ICCV}. From an intuitive perspective, these templates are constructed from a reference database by optimizing for images that minimize the average deformation to each subject within the specific population.

In this study, we use AtlasGAN \citep{Dey_2021_ICCV} to generate optimal spatiotemporal cohort-level templates with realistic anatomy. Its template generator component is illustrated in Fig. \ref{fig:pipeline}~\textbf{2)}, comprising two primary sub-networks: a template synthesis decoder and a VoxelMorph-like~\citep{DALCA2019226} registration sub-network.
The decoder learns to map the conditions (e.g., age and/or disease condition) to the residual map derived from an average scan obtained from the target database to produce the desired template. Simultaneously, the registration sub-network warps the template to instances of the target population with a smooth, invertible, and small displacement. 
Please refer to the original paper by \citet{Dey_2021_ICCV} for more details. The primary loss function of the model and important considerations when integrating this method into the InBrainSyn are discussed as follows. 

\textbf{Diffeomorphic deep learning-based registration}:
As shown in Fig. \ref{fig:pipeline} \textbf{2)}, SVF is the output of the U-Net model. 
The diffeomorphic displacement field $\phi^{(t)}$ is then determined by the following Ordinary Differential Equation (ODE),
\begin{equation}
\label{eq:ode}
    \frac{\partial \phi^{(t)}}{\partial t}=\bm v(\phi^{(t)}),
\end{equation}
with $\phi^{(0)}=Id$, representing the initial identity transformation, and $t$ representing time. The resulting diffeomorphic displacement field $\phi$ is obtained by integrating the SVF $\bm v$ over the unit time interval $t \in [0, 1]$. Due to its efficiency, the numerical integration technique of \textit{scaling and squaring} \citep{arsigny2006log, ashburner2007fast} is typically used. The integration of a stationary ODE forms a continuous one-parameter group of diffeomorphisms.
The SVF $\bm v$ can be considered as an element of the Lie algebra in the tangent space of a manifold, and the displacement field $\phi$ can be seen as a member of the Lie group in a topological manifold space. The displacement field $\phi$ is obtained through exponentiating the SVF $\bm v$ as shown in Eq.~\ref{eq:lie_exp}:
\begin{equation}
\label{eq:lie_exp}
    \exp(\bm v) \equiv \phi^{(1)} = \int_0^1 \bm v(\phi^{(t)}) \,dt.
\end{equation}
From the properties of the one-parameter subgroup, for any scalar $t$ and $t'$, the property stated in Eq.~\ref{eq:lie_property} is satisfied:
\begin{equation}
\label{eq:lie_property}
    \exp((t+t')\bm v) = \exp(t\bm v) \circ \exp(t'\bm v),
\end{equation}
where $\circ$ represents the composition map associated with the Lie group. For more details, please refer to \citet{arsigny2006log,bossa2007contributions,DALCA2019226}.

\textbf{Loss Functions:}
The generator loss function of AtlasGAN is defined as:
\begin{equation}
\label{eq:gen_loss}
    L = L_{LNCC} + \lambda_{reg}Reg(\phi) + \lambda_{GAN}L_{GAN},
\end{equation}
where $L_{LNCC}$ is the image matching similarity term, which is associated with the squared localized normalized cross-correlation objective, ensuring the standardization of the image intensity in local windows \citep{avants2011reproducible}; $\lambda_{GAN}L_{GAN}$ is the least-squares GAN term \citep{mao2017least}, which ensures the realism of the moved templates; and $\lambda_{reg}Reg(\phi)$ serves as the deformation regularization penalty, which ensures the smoothness and centrality of the displacement~\citep{DALCA2019226}. In particular, the latter comprises three terms:
\begin{equation}
\label{eq:lambda_reg}\displaystyle
    \lambda_{reg} Reg(\phi) = \lambda_1{\|\bar{u}\|}_2^2 + \lambda_2\displaystyle\Sigma_{p\in\Omega}{\|\nabla u(p)\|}_2^2 + \lambda_3\displaystyle\Sigma_{p\in\Omega}{\|u(p)\|}_2^2,
\end{equation}
where $p$ represents the voxel, $u$ represents the spatial displacement that satisfies $\phi = Id + u$
, and $\bar{u}$ denotes the moving average of $u$ in a window of $n$ updates (set to 100 in our experiments), that is, $\bar{u} = \frac{1}{n}\Sigma_{p\in\Omega}u(p)$, where $\Omega \subset \mathbb{R}^3 $ is the 3D spatial domain. The first and third terms in Eq.~\ref{eq:lambda_reg} are the magnitude terms that ensure small deformations between the template and image on both a population and individual level, respectively. The second term is the smoothness term that ensures smooth deformations, penalized with diffusion-like $L_2$ regularization, where $\lambda_{reg}$ is the vector $[\lambda_1, \lambda_2, \lambda_3]$.
 
For the discriminator, the $R_1$ gradient penalty proposed in \citet{mescheder2018training} is incorporated to enhance the stability of GAN training, and is defined as:
\begin{equation}
\label{eq:r1}
    R_1 = \frac{\lambda_{gp}}{2}\mathbb{E}_{x\sim P_{real}}[\|\nabla D(x)\|]_2^2,
\end{equation}
where $P_{real}$ indicates the real distribution, and $\nabla D(x)$ is the gradient of the discriminator for a specific input image $x$. Notably, the penalty weight $\lambda_{gp}$ plays a crucial role and requires careful tuning based on the dataset used for training.

\subsubsection{Learning Subject-Specific Aging Deformations}
As shown in Fig. \ref{fig:pipeline}~\textbf{3)}, the learned decoder and registration networks from the previous step are used to both generate cohort-specific templates and extract SVFs. 

\textbf{Template Creation Decoder}: The template synthesis network takes age $t_i$ and an average population image as inputs and generates the corresponding template $T(t_i)$ for each time point $i \in [0, ..., s]$, where $s$ represents the total number of time points.
In our work, we train independent networks for HC and AD subjects.
For simplicity, $T(t_i)$ is denoted as $T_i$, as shown in Fig. \ref{fig:pipeline} \textbf{3)}, which illustrates the example of synthesizing a time series of templates $T_0, ..., T_s$ (represented by orange cubes in the figure) given an individual MRI $I_0$.
 
\textbf{Registration U-Net}: For our purposes, to derive individual-level transformations to predict subject-specific aging from the cohort-level registration network trained for AtlasGAN, it is necessary to define both cohort-level transformations and the geodesic path between the template and the subject at equivalent ages.

To accomplish this, cohort-level transformations are initially obtained using the templates estimated in the preceding step. For instance, considering that an MRI scan of an individual, denoted as $I_0$, was acquired at age $t_0$, a series of template-to-template SVFs, represented as $\bm v_{T_{0}-T_{i}}$, can be extracted:
\begin{equation}
\label{eq:regis_T2F}
\bm g_{\theta}(T_i, T_0)=\bm v_{T_0-I_i},
\end{equation}
where $\theta$ represents the parameters of the \textit{learned registration network} denoted as $\bm g$.
These SVFs correspond to the registration between $T_0$ and $T_i$, with $i$ ranging from 1 to $s$ to denote various target time points, as illustrated in the bottom right corner of Fig. \ref{fig:pipeline} (indicated by the purple cubes). 

Next, the template-subject correspondence is estimated between $T_0$ and the subject MRI $I_0$, represented as $\bm v_{T_{0}-I_{0}}$.
  
\textbf{Parallel Transport with the Pole Ladder}:
Following the extraction of a series of template-to-template SVFs and a template-to-subject SVF, the pole ladder \citep{lorenzi2014efficient} is employed for obtaining the trajectory of individual morphological evolution.

We will briefly outline the procedure for transporting a single deformation from the template space to the individual space using the pole ladder and note that this method can be readily extended to scenarios involving multiple deformations. Additionally, we emphasize the compatibility of the pole ladder with a DL-based diffeomorphic registration setting.

\begin{figure}[t]
    \centering
    \includegraphics[width=0.48\textwidth]{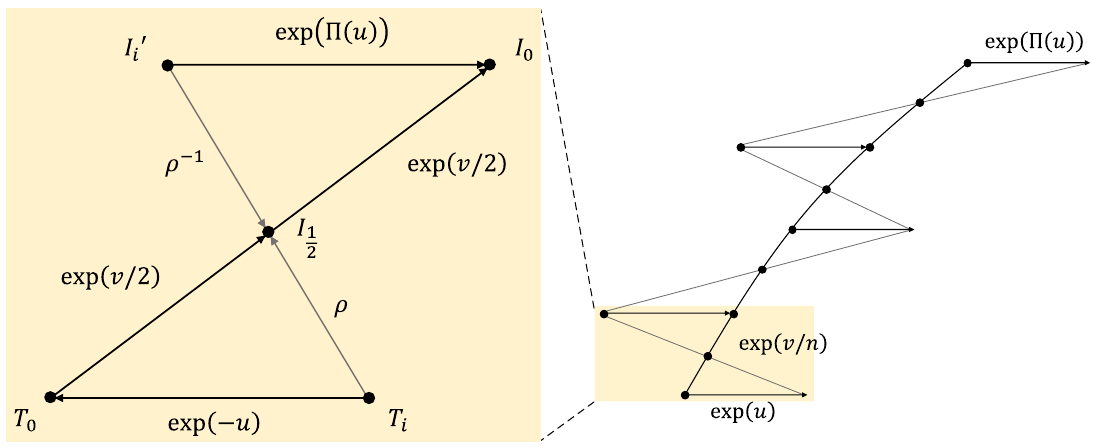}
    \caption{Illustration of the pole ladder adapted from \citet{lorenzi2014efficient} for transporting a single deformation between templates $T_0$ and $T_i$ to estimate the stationary velocity field (SVF) connecting the reference individual image $I_0$ and the predicted individual image $I_i'$. Steps include obtaining longitudinal SVF between templates ($\bm u$), SVF between template and individual image ($\bm v$), and deriving diffeomorphisms for parallel transport. The resulting parallel transported deformation is obtained through the conjugate actions of diffeomorphisms parameterized by SVFs $\bm u$ and $\bm v$. A truncated Baker-Campbell-\reviadd{Hausdorff} (BCH) formula is used for computations (denotes as $\Pi$), and a scaling factor $n$ ensures computational efficiency.}
    \label{fig:PT}
\end{figure}

To provide a straightforward illustration of the pole ladder's application for transporting a single deformation, let's consider two templates as shown in the left of Fig.~\ref{fig:PT}: a reference $T_0$ and a target $T_i$. $T_0$ is chosen as the template that matches the age of the given MRI scan $I_0$. 
Thus, the goal of the pole ladder is to parallel transport the deformation between $T_0$ and $T_i$ to estimate the SVF connecting $I_0$ and $I_i'$  within the individual space. This SVF is used to obtain the diffeomorphic deformation field $\phi$ that is used to compute $I_i'$ as $I_i' = \phi \ast I_0$, where $\ast$ represents the warping operator. $I_i'$ represents the appearance of the given MRI in $t_i-t_0$ years ahead (or backwards if it is negative).
 
As introduced in Section~\ref{sec:atlasgan}, these diffeomorphisms can be parameterized by SVFs through Lie group exponentials. Consequently, geodesic paths within the space of diffeomorphisms can be defined using such one-parameter subgroup parametrization, enabling the definition of corresponding geodesics within the space of images. In this manner, for each template deformation from deformations set $\{\bm{v}_{T_0 - T_i} \mid i = 1, \ldots, s\}$, the implementation steps for the parallel transport of longitudinal deformation can be summarized as follows:

\begin{enumerate}
  \item Take the longitudinal SVF $\bm u = \bm v_{T_{0}-T_{i}}$ between the template $T_0$ and the template $T_i$ such that $T_i = \exp(\bm u) \ast T_0$.
  
  \item Take the SVF $\bm v=\bm v_{T_{0}-I_{0}}$ between the template $T_0$ and the reference individual image $I_0$. The half-space image is given by $I_{\frac{1}{2}} = \exp(\bm v/2) \ast T_0 = \exp(-\bm v/2) \ast I_0$.
  
  \item The diffeomorphism from the template $T_i$ to the half-space image $I_{\frac{1}{2}}$ is denoted as $\rho = \exp(\bm v/2) \circ \exp(-\bm u)$, and thus, the diffeomorphism from the reference individual image $I_0$ to the predicted individual image $I_i'$, i.e., the parallel transported longitudinal deformation is represented as Eq.~\ref{eq:pt_deformation}:
  \begin{equation}
    \label{eq:pt_deformation}
        \exp(\Pi(\bm u, \bm v)) = \exp(\bm v/2) \circ \rho^{-1} = \exp(\bm v/2) \circ \exp(\bm u) \circ \exp(-\bm v/2).
  \end{equation}
\end{enumerate}

As outlined in Eq. \ref{eq:pt_deformation}, the resulting parallel transported deformation, denoted as $\exp(\Pi(\bm u))$, is achieved through the conjugate actions associated with the diffeomorphism parameterized by $\bm v/2$ and $\bm u$. In practice, a truncated Baker-Campbell-\reviadd{Hausdorff} (BCH) formula, as proposed in \citet{bossa2007contributions}, is employed to approximate the logarithm of the composition of the diffeomorphisms, working within the Lie algebra. Additionally, when dealing with non-infinitesimal diffeomorphisms during the parallel transport process, computations need to be executed recursively to construct a successive ladder. To accommodate this, a scaling factor denoted as $n$ is introduced to ensure that $\bm v/n$ remains sufficiently small, as shown in the right part of Fig.~\ref{fig:PT}. The pole ladder algorithm utilizing the BCH formula is further detailed in Algorithm \ref{alg:pl_bch}. 

\begin{algorithm}[t]
\caption{Parallel transport of SVFs with the pole ladder \citep{lorenzi2014efficient}}
\label{alg:pl_bch}
\KwIn{$\bm u$: the template-to-template SVF, \\
\hspace*{3.1em}$\bm v$: the template-to-subject SVF}

\KwOut{$\Pi_{BCH}(\bm u,\bm v)$: the parallel transported SVF $\bm u$ along the SVF $\bm v$}
\BlankLine
\BlankLine
\tcp{{Scaling step:} compute $n$ such that $\bm v/n$ is smaller than the voxel size in all dimensions} 
\BlankLine
    $n = \left\lceil\frac{{max}_{p \in \Omega} \|\bm v(p)\|}{ voxelsize}\right\rceil$
\tcp*{$\lceil \cdot \rceil$ is the ceiling operation}
\BlankLine
\BlankLine
\tcp{Parallel transporting step using the Baker-Campbell-\reviadd{Hausdorff} (BCH) formula} 
\BlankLine
$\bm u_0=\bm u$\;

\For{$j \leftarrow 1$ \KwTo $n$}{
    {$\bm u_{j} = \bm u_{j-1} + \left[\frac{\bm v}{2n}, \bm u_{j-1}\right] + \frac{1}{2}\left[\frac{\bm v}{2n}, \left[\frac{\bm v}{2n}, \bm u_{j-1}\right]\right]$}	\tcp*{$[,]$ is the Lie bracket} 
}
\BlankLine
$\Pi_{BCH}(\bm u,\bm v) = \bm u_n$\;
\BlankLine
\BlankLine
\end{algorithm}

Hence, by leveraging the pole ladder, the series of extracted longitudinal template-to-template SVFs is parallel transported into an individualized series of subject-to-subject SVFs as shown in Fig. \ref{fig:pipeline} \textbf{3)} (represented by red cubes).

\subsubsection{Synthesizing Individual Time Series Images}
The final subject-specific synthesis is achieved by individualized transformations as illustrated in Fig. \ref{fig:pipeline} \textbf{4)}. First, the individualized SVFs are integrated to yield a series of longitudinal displacement fields. Second, these displacement fields are then applied to warp the reference individual scan. This pipeline ensures that the predicted scans possess both anatomical plausibility via diffeomorphic deformations and personalized features via parallel transport.

We investigated two scenarios of longitudinal synthesis. First, we synthesize scans at new time points that align with an individual's current health status. These predictions can follow either a normal aging trajectory or an AD progression trajectory; In this case, all templates $T_0 ... T_s$ come from the same cohort.
Second, for subjects undergoing a transition from health to disease, we use first $T_0 ... T_k$ templates from the healthy cohort and then $T_{k+1} ... T_s$ from the disease cohort. This way, we can simulate the onset of AD at a specific time point $k+1$. 

\section{Results}\label{sect:results}
\subsection{Dataset}
We use the OASIS version 3 (OASIS-3) dataset \citep{lamontagne2019oasis} as the reference database. OASIS-3 is a longitudinal multimodal neuroimaging, clinical, cognitive, and biomarker dataset for normal aging and AD\footnote{\url{https://www.oasis-brains.org/##about}}. OASIS-3 has data from 1,378 participants, including 755 cognitively normal adults and 622 individuals at various stages of cognitive decline ranging from 42 to 95 years old. It contains over 2,000 magnetic resonance (MR) sessions and includes T1w MR scans, among other sequences. \reviadd{The data was collected using different Siemens scanners, including both 1.5T and 3T. More information on the dataset can be found in the OASIS-3 image data dictionary\footnote{\url{https://sites.wustl.edu/oasisbrains/files/2024/04/OASIS-3_Imaging_Data_Dictionary\_v2.3-a93c947a586e7367.pdf}}.} In this study, we use T1w scans. 

\begin{table}[!t]
    \begin{center}
    \caption{A summary of the original OASIS-3 dataset and the number of subjects and scans included in our study.}
    \label{tab:dataset}
    \begin{tabular}{ccc}
    \toprule
    {} & \textbf{\# subjects} & \textbf{\# T1w scans}\\
    \midrule
    Collected & 1,316 & 2,681\\
    \addlinespace
    \begin{tabular}[c]{@{}c@{}}
        Excluded\\
    \end{tabular} & 1 & 3\\
    \addlinespace
    \begin{tabular}[c]{@{}c@{}}
        Remained\\
        (HC/AD/non-AD)
    \end{tabular} & 
    \begin{tabular}[c]{@{}c@{}}
        1,315\\
        (739/419/157)
    \end{tabular} &
    \begin{tabular}[c]{@{}c@{}}
        2,678\\
        (1,678/688/312)
    \end{tabular}\\
    \addlinespace
    \midrule
    \begin{tabular}[c]{@{}c@{}}
        \textbf{Included in this study}\\
        \textbf{(HC/AD)}
    \end{tabular} &
    \begin{tabular}[c]{@{}c@{}}
        1,158\\
        (739/419)
    \end{tabular} &
    \begin{tabular}[c]{@{}c@{}}
        2,366\\
        (1,678/688)
    \end{tabular}\\
    \bottomrule
    \end{tabular}
    \end{center}
\end{table}

\textbf{Data Preparation}:
We collected the FreeSurfer processed OASIS-3 dataset released in July 2022. 
The summary of this collection is shown in Table~\ref{tab:dataset}, in which the number of subjects and scans are counted separately. There were 2,681 successfully collected scans, in which two scans failed the FreeSurfer quality control (QC) procedure (i.e., marked as \emph{Quarantined}) and one scan failed to define clinical dementia rating (CDR) score, which, in our application, is relevant when considering transition cases. Therefore, they were discarded for the study. We also filtered the remaining scans according to the diagnoses provided in OASIS-3 and further removed non-AD dementia types to keep the focus on AD progression in our study. \reviadd{For this, we only chose subjects with “AD” or “DAT”  (Dementia of the Alzheimer's Type) as the diagnosis provided by the dataset.} In this cohort, the clinical stage was defined by CDR following standards, as follows: a CDR of 0 corresponds to normal cognitive function, CDR = 0.5 indicates very mild cognitive impairment, CDR = 1 indicates mild dementia, CDR = 2 indicates moderate dementia, and CDR = 3 indicates severe dementia. According to OASIS-3's acquisition protocol, participants who reached CDR = 2 were no longer eligible for in-person assessments. For this reason, subjects with CDR = 2 are scarce in this dataset and CDR=3 are non-existent. 

After curation, 2,366 scans from 1,158 subjects were included in this study. We define HC subjects as those with CDR = 0 at all visits. 1,678 scans from 739 subjects complied with this requirement. In turn, the AD cohort is composed of individuals who progressed to clinical AD dementia at some point during follow-up visits but could have CDR$>=$ 0 at the earlier MRI scanning time. While some scans of the AD group can have CDR = 0, they are not mixed with the HC group since these images might already show some signs of AD progression that were not yet clinically signaled by the CDR test. Thus, we used 688 images from 419 subjects for the AD group with CDR scores from 0 to 2. Approximately 80$\%$ of the data was allocated to the training set, while the remaining 20$\%$ was assigned to the test set. Only the test set was used for the results reported in Sect. \ref{sec:eva_cohort}, while the whole database was used for the results reported in Sect. \ref{sec:eva_individual}.
 
\textbf{Neuroimaging Processing}:
We followed a similar preprocessing protocol to \citet{Dey_2021_ICCV}. Specifically, we used T1w data preprocessed with FreeSurfer (i.e., \emph{norm.mgz}). \reviadd{These images have an isotropic resolution of 1mm}. FreeSurfer performs skull-stripping and bias field correction according to the FreeSurfer process flow\footnote{\url{https://surfer.nmr.mgh.harvard.edu/fswiki/ReconAllDevTable}}. Then, affine registration was applied to the image using the FreeSurfer \emph{mri\_vol2vol} command, utilizing Talairach space encoded in \emph{talairach.xfm} to Montreal Neurological Institute (MNI) 305 space. The segmentation masks for each image can be obtained by SynthSeg \citep{billot_synthseg_2023} for assessment purposes. 
\reviadd{We then rescaled the image intensity range to [0,1] and cropped the input scan to [208, 176, 160].} This grid size was calculated to preserve most foreground information using 200 randomly selected images from the dataset.

\begin{figure*}[t]
    \centering
    \includegraphics[width=0.8\textwidth]{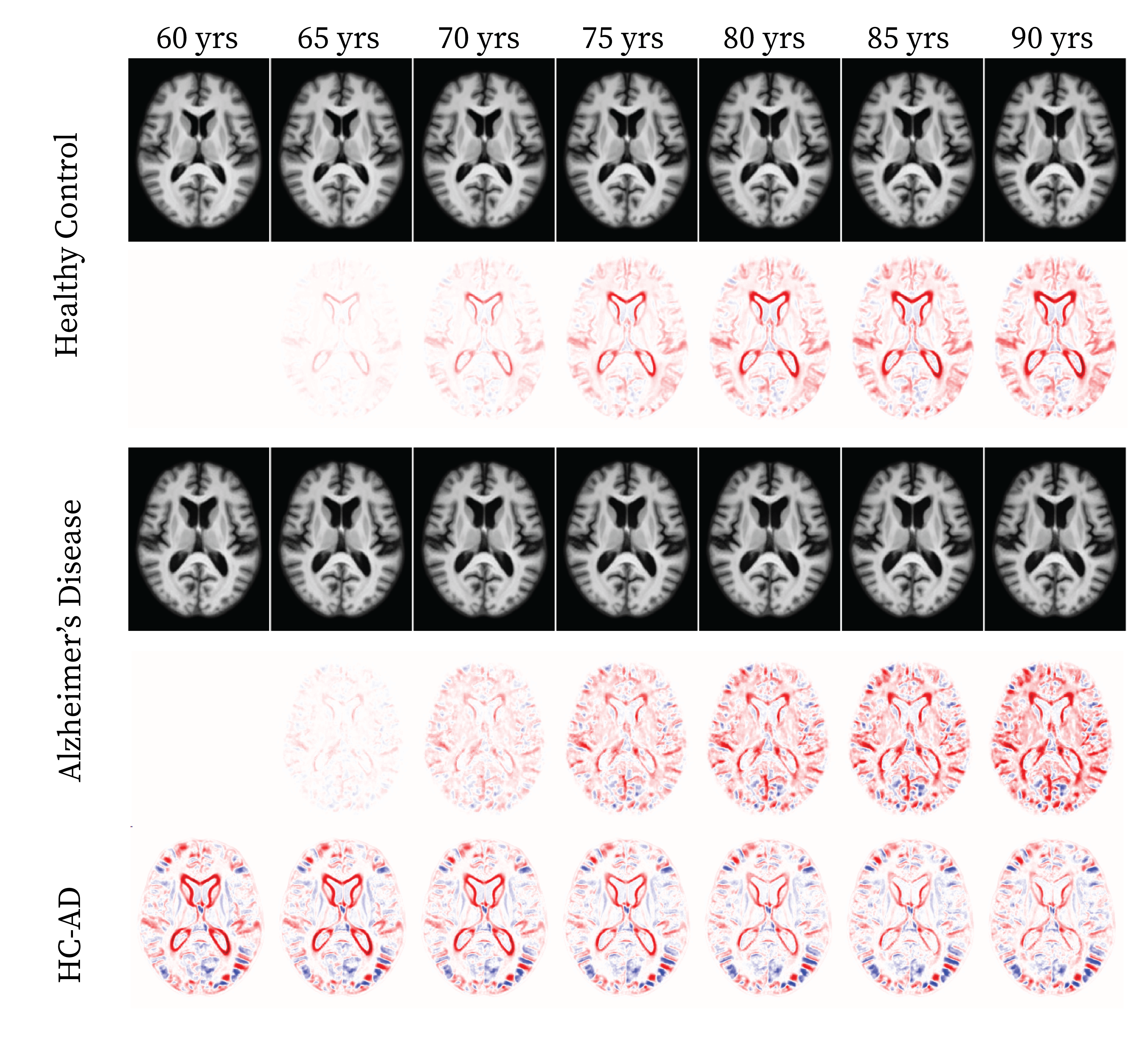}
    \caption{Synthetic deformable templates for Healthy Control (HC) and Alzheimer's Disease (AD) cohorts from 60 to 90 years old with 5-year intervals from the axial plane at the 80th slice (the first and the third rows, respectively). The intensity difference maps for each cohort are calculated between the corresponding synthetic template and the one at age 60 (the second and the fourth rows, respectively). The last row shows the difference between HC and AD cohorts at matched ages.}
    \label{fig:cohort_templates}
\end{figure*}

\begin{figure}[t]
    \centering
    \includegraphics[width=0.49\textwidth]{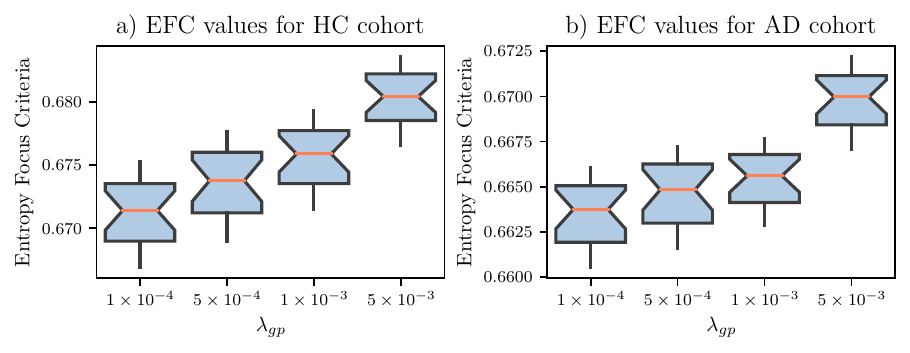}
    \caption{Entropy focus criteria (EFC) values of synthetic templates from integer age ranging from 60 to 80, using different $\lambda_{gp}$ for healthy control (HC) and Alzheimer's disease (AD) cohorts. Lower EFC values indicate sharper images.}
    \label{fig:efc_for_cohorts}
\end{figure}
 
\begin{figure}[t]
    \centering
    \includegraphics[width=0.49\textwidth]{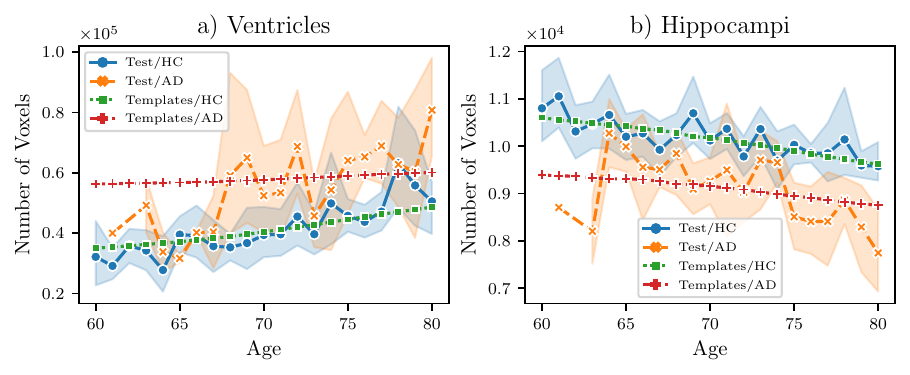} 
    \caption{Volumetric trends of synthetic template segmentations for healthy control (HC) and Alzheimer's disease (AD) cohorts overlaid upon the volumetric trends for the underlying HC (blue) and AD (orange) test sets. }
    \label{fig:volumetric_trends}
\end{figure}

\textbf{Hyperparameter Selection}:
In this study, following the hyperparameters adopted in the previous works of \citet{Dey_2021_ICCV, dalca2019learning, mescheder2018training}, the hyperparameters were set as follows:
\begin{itemize}
    \item $\lambda_{GAN} = 0.1$,
    \item $\lambda_{reg} = [\lambda_1, \lambda_2, \lambda_3] = [1, 1, 0.01]$, 
\end{itemize}
Regarding $\lambda_{gp}$, we tested four values: $10^{-4}, 5\times10^{-4}, 10^{-3}, \mathrm{ and } \ 5\times10^{-3}$.
These four values were selected based on previous experience with neuroimaging datasets. Subsequently, the weight yielding the best sharpness of the synthetic templates for the trained AtlasGAN was chosen for further experiments.

\subsection{Evaluation of Template Generation}
\label{sec:eva_cohort}

Three assessments were performed on template synthesis in terms of its quality and accuracy in representing HC and AD cohorts: 1) visual inspection of learned templates; 2) Entropy Focus Criterion (EFC) \citep{atkinson1997automatic}: to evaluate the sharpness of the synthetic templates and determine the best model used for sequencing steps; 3) Volumetric trends: to compare the volumes of anatomical structures within the synthesized templates to underlying real distribution by comparing segmentation-derived volumes.  
 
Firstly, the results of template generation are visually presented in Fig. \ref{fig:cohort_templates}, where we show the synthesized deformable templates from 60 to 80 years old with a 5-year interval for the HC and AD cohorts. Specifically, the second row and fourth row of Fig. \ref{fig:cohort_templates} illustrate the residual maps of the generated templates, depicting the deviations between each subsequent template and the template corresponding to the age of $60$ years old. These residual maps were derived by subtracting the $60$-year-old template from the subsequent templates. We also show the difference maps between the HC and AD templates at equivalent ages in the last row.

Secondly, the individual template quality is assessed via EFC. AtlasGAN models trained with four different $R_1$ gradient penalty weights $\lambda_{gp}$ of the discriminator are obtained for each cohort. We used the trained decoder to synthesize the templates from integer age from 60 to 80 for each model. The results are shown in Fig. \ref{fig:efc_for_cohorts} for the two cohorts, the model employing a $\lambda_{gp}$ value of $10^{-4}$ exhibit the smallest EFC values for the learned templates. Therefore, we choose this model to evaluate the subsequent experiments.  

Thirdly, it is also necessary to verify whether the templates correctly capture the morphological changes in the two cohorts; thus, we use a segmentation-based volumetric analysis. SynthSeg \citep{billot_synthseg_2023} was employed to get the segmentation masks for real MRI scans from the database and the synthetic templates. In particular, two specific brain regions—the ventricles and the hippocampi—are of greater interest in normal aging or AD studies. The results of the ventricles and hippocampi volumetric count trend are shown in Fig. \ref{fig:volumetric_trends}. 

\subsection{Evaluation of Individual-Level Generation}
\label{sec:eva_individual}

\begin{figure*}[!t]
    \centering
    \includegraphics[width=\textwidth]{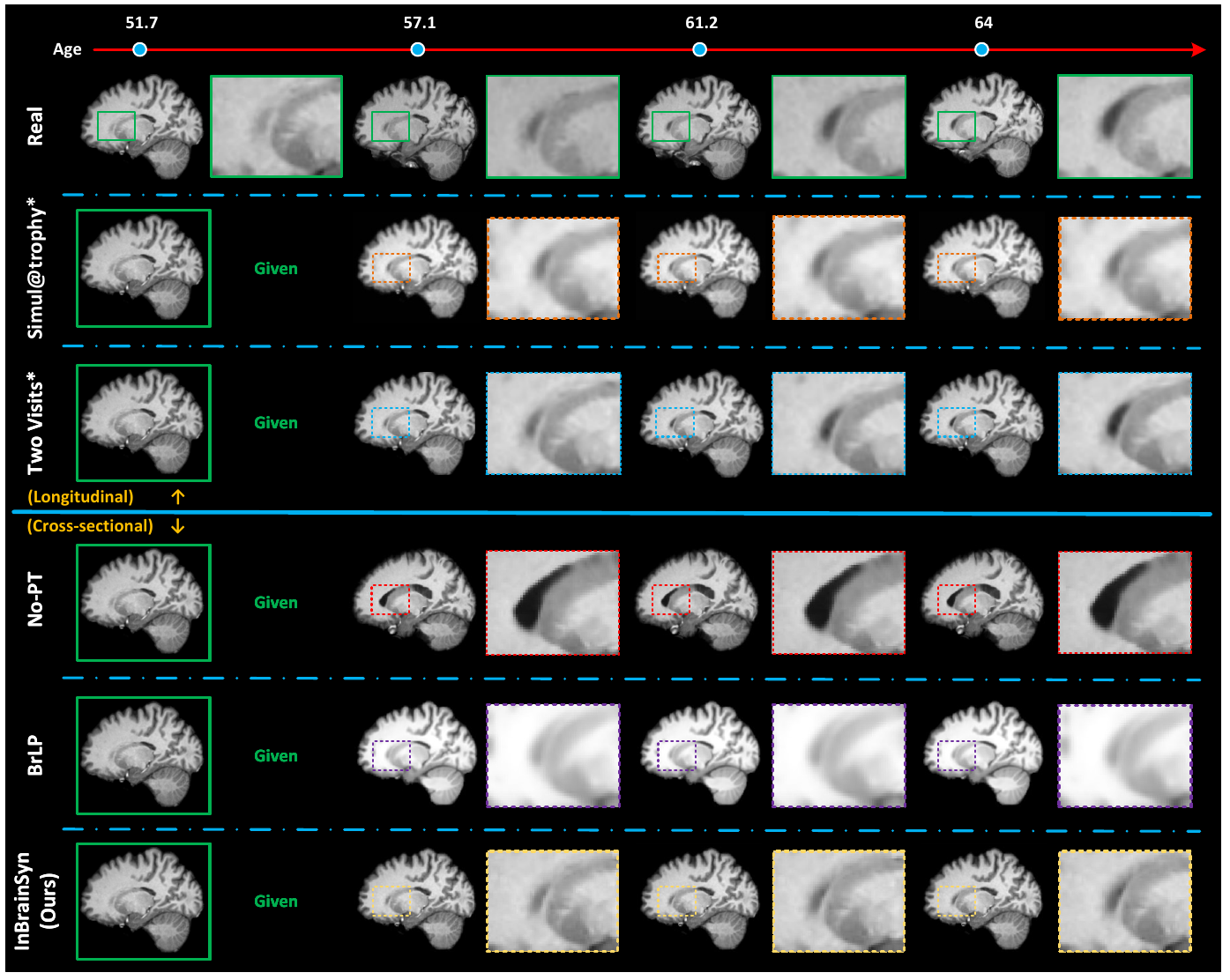}
    \caption{\textbf{Qualitative comparison}: the synthetic MRI scans for a longitudinal representative subject \textit{OASIS31167} across the No-PT and \reviadd{three} brain MRI simulators. The real scan is enclosed within the green solid box, while the synthetic scans are enclosed within the dashed box. \reviadd{Methods requiring longitudinal data or real longitudinal segmentation maps are marked with an asterisk (*). All other competing methods rely on a single input for prediction. (For better visualization, please refer to the online version.)}}
    \label{fig:Individual_comparisons}
\end{figure*}

\begin{figure}[!ht]
    \centering
    \includegraphics[width=0.49\textwidth]{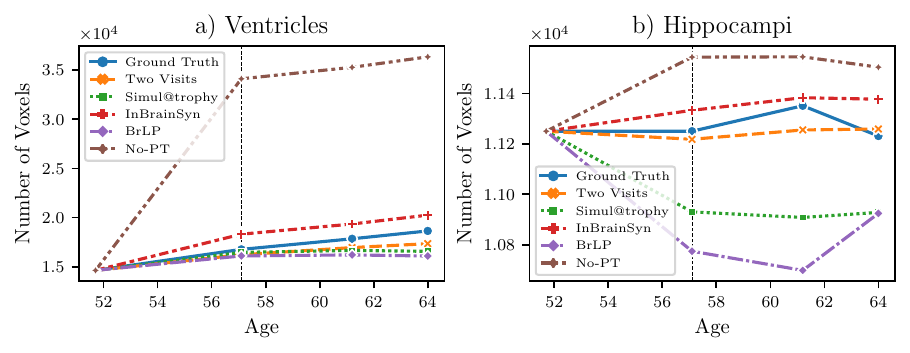}
    \caption{\reviadd{T}he volumetric trends on ventricles and hippocampi regions for synthetic MRI scans of a longitudinal representative subject \textit{OASIS31167} across the No-PT and \reviadd{four} brain MRI simulators. \reviadd{The ground truth is depicted as a solid blue line for reference.}}
    \label{fig:Individual_trends_comparison}
\end{figure}

Assessments of individual-level generation focus on two aspects: the quality of synthetic MRI scans and the accuracy of modeling individual characteristics. For assessing the quality of synthetic MRI scans, six distinct and widely recognized similarity metrics were employed following \citet{fu2023fast}: mean absolute error (MAE), structural similarity index (SSIM), normalized cross-correlation (NCC), peak signal-to-noise ratio (PSNR), normalized Frobenius norm (NFN), and Dice score (DSC).
For assessing the accuracy of modeling individual characteristics, \reviadd{we compute the mean absolute error (MAE) between the volumes of actual follow-up scans and the generated brain MRIs across five subcortical structures—hippocampi, amygdalae, thalami, caudates, and putamina—as well as ventricles and cerebrospinal fluid (CSF) regions. 
The MAE between the synthetic scans and the ground truth scans is estimated as:}
\reviadd{
\begin{equation}
\label{eq:MAE}
MAE_{r} =\bigg | \frac{Vol(I_s, r)}{Vol(I_s, wb)} - \frac{Vol(I_s^*, r)}{Vol(I_s^*, wb)} \bigg | * 100
\end{equation}
}
\reviadd{where $I_s$ represents the ground truth scan for subject $s$, $I_s^*$ is the simulated scan, $Vol(I_s, r)$ denotes the estimated regional volumes in $mm^3$ for structure $r$, and $wb$ refers to the whole brain. We also visually inspect individual time-series MRI scans, compared with baselines.}

\begin{table*}[!t]
\renewcommand{\arraystretch}{1.5}
\centering
\caption{\reviadd{Mean absolute error (MAE) (mean $\pm$ standard deviation) between predicted and ground truth MRI scans across brain regions, expressed as a percentage of total brain volume. 
Methods requiring longitudinal data or real longitudinal segmentation maps are shown as a reference and marked with an asterisk (*). The best results of methods requiring a single image are highlighted in \textbf{bold}.}}
\label{tab:MAE}
\resizebox{\textwidth}{!}{%
\begin{tabular}{cc|ccccccc}
\toprule
\multirow{2}{*}{\textbf{Cohort}} & \multirow{2}{*}{\textbf{Method}} & \multicolumn{7}{c}{\textbf{Regions} (MAE \% ($\downarrow$))} \\ \cline{3-9} 
                        &   & \textbf{Ventricles} & \textbf{Hippocampi} & \textbf{Amygdalae} & \textbf{Thalami} & \textbf{Caudates} & \textbf{Putamina} & \textbf{CSF} \\ 
\midrule
\multirow{5}{*}{\shortstack{\textbf{HC} \\ ($n=868$)}}    
 & Simul@trophy$^*$        & 0.329 (0.320)& 0.013 (0.011)& 0.010 (0.009)& 0.034 (0.025)& 0.015 (0.014)& 0.014 (0.013)& 0.569 (0.439)\\
 & Two Visits$^*$          & 0.177 (0.122)& 0.017 (0.015)& 0.012 (0.011& 0.026 (0.022)& 0.022 (0.021)& 0.019 (0.016)& 0.778 (0.718)\\  \cline{2-9}
 & No-PT& 0.622 (0.503)& 0.035 (0.027)& 0.019 (0.015)& 0.048 (0.036)& 0.071 (0.046)& 0.077 (0.047)& 1.669 (0.924)\\
& BrLP                    & 0.316 (0.333)& 0.021 (0.016)& 0.015 (0.011)& 0.041 (0.031)& 0.021 (0.016)& 0.031 (0.023)& 0.859 (0.671)\\
 & InBrainSyn (Ours)             & \textbf{0.204 (0.238)}& \textbf{0.014 (0.011)}& \textbf{0.008 (0.006)}& \textbf{0.019 (0.017)}& \textbf{0.018 (0.018)}& \textbf{0.020 (0.014)}& \textbf{0.636 (0.662)}\\
 \midrule
\multirow{5}{*}{\shortstack{\textbf{AD} \\ ($n=184$)}}
  & Simul@trophy$^*$      & 0.498 (0.480)& 0.018 (0.017)& 0.011 (0.009)& 0.041 (0.037)& 0.018 (0.018)& 0.018 (0.017)& 0.538 (0.433)\\
 & Two Visits$^*$          & 0.256 (0.143)& 0.022 (0.018)& 0.011 (0.011)& 0.027 (0.019)& 0.024 (0.020)& 0.023 (0.021)& 0.834 (0.756)\\ \cline{2-9}
 & No-PT                & 1.126 (0.784)& 0.046 (0.036)& 0.021 (0.017)& 0.053 (0.038)& 0.060 (0.039)& 0.070  (0.043)& 2.057 (1.070)\\
 & BrLP                    & 0.544 (0.511)& 0.033 (0.025)& 0.021 (0.014)& 0.049 (0.039)& \textbf{0.021 (0.018)}& 0.041 (0.027)& 1.265 (0.767)\\
 & InBrainSyn (Ours)             & \textbf{0.433 (0.459)}& \textbf{0.026 (0.021)}& \textbf{0.016 (0.010)}& \textbf{0.049 (0.033)}& 0.031 (0.021)& \textbf{0.028 (0.020)}& \textbf{0.681 (0.681)}\\
\bottomrule
\end{tabular}
}
\end{table*}

\reviadd{\textbf{Baselines}: We compare InBrainSyn with two methods: }\reviiadd{a no parallel transport (No-PT) baseline and Brain Latent Progression (BrLP) \citep{puglisi2024enhancing,puglisi2025}.  
The No-PT is achieved by registering the given scans to cohort-level templates without acquiring individual-level transformations (i.e., without parallel transport). This setting serves as an ablation study to highlight the importance of accounting for inter-subject differences in the generation process.}
\reviadd{BrLP is a spatiotemporal disease progression model based on latent diffusion. BrLP has been reported to outperform previous learning-based methods \citep{puglisi2025}, such as CounterSynth \citep{pombo2023equitable}, DANI-Net \citep{ravi2022degenerative} and Latent-SADM \citep{yoon2023}. We utilize the pre-trained models from the original study to generate results. Since BrLP synthesizes MRI scans at 1.5 \text{mm$^3$} resolution; we apply linear interpolation to obtain 1 \text{mm$^3$} resolution to match the resolution of real data.}

\reviadd{In addition to these two methods, we compare the results with two reference methods that require longitudinal data: Simul@trophy and Two Visits. These methods can be seen as upper bounds for the methods that use a single MRI volume as an input. Simul@trophy uses a biomechanical approach to generate the images \citep{khanal2017simulating}. In addition to the input image, Simul@trophy requires a predefined atrophy map. In our experiments, we estimate such atrophy maps to guide the generation process using actual longitudinal segmentation maps of CSF and brain parenchyma. We down-sample the MRI resolution (by a factor of 2) to achieve computationally feasible reference times and fit in the memory. Subsequently, the resolution was restored to its original state using linear interpolation. Two Visits is a registration-based method we introduced in our previous work \citep{fu2023fast}. It uses diffeomorphic registration to interpolate between two MRI scans of the same individual. Since it relies on both an initial and a final time point, it requires longitudinal data.}

\subsubsection{Evaluation \reviadd{on HC and AD Cohorts}}
\label{sec:eva_intra_cohort}

\textbf{Qualitative comparison:}
\reviadd{We present an example subject in Fig.~\ref{fig:Individual_comparisons}, comparing the generated scans among methods with real longitudinal ground truth data as reference. The first row displays the real ground truth images alongside zoomed-in regions to aid visualization.}
\reviadd{InBrainSyn generates the most realistic images compared to the No-PT and BrLP. These images are also sharper than the ones generated with Simul@trophy. Two Visits is the method that generates the most accurate images, which is expected as it requires longitudinal data}. 

In Fig. \ref{fig:Individual_trends_comparison}, we present the volumetric trends for two specific brain structures across the No-PT and \reviadd{four} brain MRI simulators, along with the ground truth scans \reviadd{for a subject with four real scans in OASIS-3}. Notice that all three methods start from the same initial scan, which is the first available scan at age 51.7. \reviadd{As shown, InBrainSyn yields better results than other methods that require a single image, especially in the hippocampi, and is competitive with the Two Visits method.}

\begin{table*}[t]
\renewcommand{\arraystretch}{1.5}
  \begin{center}
    \caption{\reviadd{Evaluation of six similarity metrics (mean $\pm$ standard deviation) for cohort-level longitudinal predictions.
    Methods requiring longitudinal data or real longitudinal segmentation maps are shown as a reference and marked with an asterisk (*). The best results of the methods requiring a single image are highlighted in \textbf{bold}.}}
    \label{tab:longi_cohort}
    \makebox[\textwidth]{%
    \begin{tabular}{cc|cccccc}
    \toprule
    \textbf{Cohort} & \textbf{Method} & \textbf{NFN} ($\downarrow$) & \textbf{MAE} ($\downarrow$)& \textbf{PSNR} ($\uparrow$)& \textbf{SSIM} ($\uparrow$)& \textbf{NCC} ($\uparrow$)& \textbf{DSC} ($\uparrow$)\\
    \midrule
    \multirow{5}{*}{\shortstack{\textbf{HC} \\ ($n=868$)}} 
     & Simul@trophy$^*$ & 0.064 (0.024)& 0.031 (0.014)& 24.45 (3.10)& 0.889 (0.057)& 0.971 (0.021)& 0.811 (0.075)\\
     & Two Visits$^*$    & 0.052 (0.023)& 0.025 (0.013)& 26.53 (3.63)& 0.936 (0.043)& 0.985 (0.014)& 0.862 (0.045)\\\cline{2-8}
     & No-PT      & 0.087 (0.016)& 0.040 (0.010)& 21.36 (1.45)& 0.800 (0.025)& 0.937 (0.012)& 0.730 (0.031)\\
     & BrLP & 0.076 (0.016)& 0.036 (0.010)& 22.51 (1.73)& 0.848 (0.036)& 0.958 (0.012)& 0.832 (0.025)\\
     & InBrainSyn (Ours)    & \textbf{0.061 (0.025)}& \textbf{0.029 (0.014)}& \textbf{24.97 (3.31)}& \textbf{0.903 (0.055)}& \textbf{0.974 (0.020)}& \textbf{0.851 (0.060)}\\
    \midrule
    \multirow{5}{*}{\shortstack{\textbf{AD} \\ ($n=184$)}}
     & Simul@trophy$^*$ & 0.062 (0.024)& 0.030 (0.013)& 24.74 (3.31)& 0.893 (0.055)& 0.968 (0.022)& 0.809 (0.079)\\
     & Two Visits$^*$    & 0.048 (0.021)& 0.023 (0.011) & 27.07 (3.59)& 0.944 (0.032)& 0.986 (0.011)& 0.865 (0.035)\\\cline{2-8}
     & No-PT & 0.087 (0.015)& 0.039 (0.008)& 21.35 (1.40)& 0.802 (0.018)& 0.929 (0.013)& 0.718 (0.040)\\
     & BrLP & 0.077 (0.016)& 0.036 (0.009)& 22.44 (1.73)& 0.852 (0.029)& 0.954 (0.015)& 0.817 (0.041)\\
     & InBrainSyn (Ours)    & \textbf{0.061 (0.022)}& \textbf{0.028 (0.012)}& \textbf{24.83 (3.00)}& \textbf{0.901 (0.044)}& \textbf{0.970 (0.018)}& \textbf{0.828 (0.046)}\\
    \bottomrule
    \end{tabular}
    }
  \end{center}
\end{table*}

\reviadd{\textbf{Quantitative comparison}:} \reviadd{Table \ref{tab:MAE} and  \ref{tab:longi_cohort} 
show the results on a large sample size ($n = 1,052$\footnote{Includes all follow-up scans with at least a one-year age difference per subject in the dataset.}) from OASIS-3. In particular, Table \ref{tab:MAE} presents MAE of regional brain volumes across seven regions in both cohorts. Notably, Simul@trophy and Two Visits achieve the best or second-best performance for many regions, benefiting from exposure to longitudinal images or segmentation maps. However, in the more realistic scenario of only a single observation being available, InBrainSyn outperforms other tested single-observation methods.}

The results of six similarity metrics are summarized in Table \ref{tab:longi_cohort}.\reviadd{ We observe that Two Visits and Simul@trophy, both of which leverage longitudinal data, achieve strong performance across all metrics. However, Simul@trophy performs worse, likely due to the downsampling effect required for computational feasibility. In contrast, our method circumvents this limitation by operating directly on vector fields. Meanwhile, our proposed InBrainSyn shows competitive image-level similarity while achieving the highest DSC accuracy among methods that rely on a single observation.} 

\subsubsection{\reviadd{Evaluation on Disease Transition Cases}}

\label{sec:eva_inter_cohort}
\begin{figure*}[!ht]
    \centering    \includegraphics[width=\textwidth]{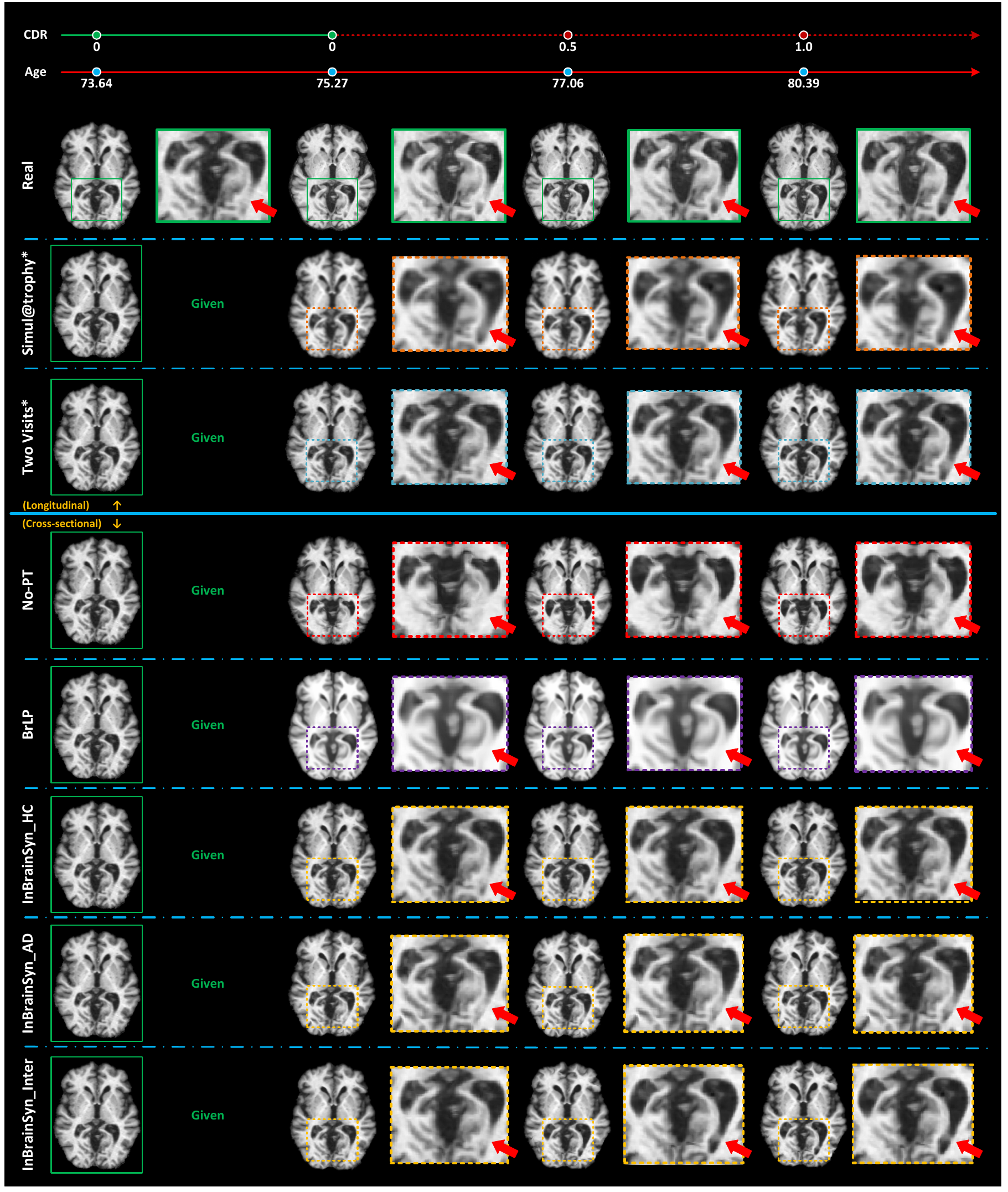}
    \caption{\textbf{Qualitative comparison}: the synthetic MRI scans for a longitudinal representative subject \textit{OASIS30331} \reviadd{using different brain MRI simulators, including} three variants of our approach for simulating AD transition case. In CDR, the green solid line indicates healthy evolution, while the red dashed line indicates AD evolution. \reviadd{A ROI of the real scans are enclosed within green solid boxes, while the synthetic scans are enclosed within the dashed boxes with different colors. Methods requiring longitudinal data or real longitudinal segmentation maps are marked with an asterisk (*). All other competing methods rely on a single input for prediction. We mark a ROI with red arrows to visually highlight the age evolution in this subject. (For better visualization, please
refer to the online version.)} }
    \label{fig:indiv_compa_331}
\end{figure*}

AD is characterized as a progressive disease, signifying that a subject may transition from healthy to cognitively unimpaired to mild cognitive impairment and to AD dementia stages \citep{kelley2007alzheimer}. Simulating the evolution of a subject, especially transitioning from cognitive normal to AD dementia, involves determining the age at which this transition occurs. To evaluate disease transition cases, we select subjects based on the following criteria: i) the number of scans per subject should be greater than 2; ii) the sequence should start with an HC state (CDR=0) and conclude with an AD state (CDR$>=$1). Only three subjects meet these specific requirements in the OASIS-3 dataset: \textit{OAS30331}, \textit{OAS30869} and \textit{OAS30899}. \reviadd{We evaluate the generated images by comparing them to ground truth data across three variants of our method and three other baselines. To ensure a fair comparison, we use CDR to guide the generation of BrLP, as it allows for disease conditioning.}

\begin{table*}[t]
\renewcommand{\arraystretch}{1.5}
\centering 
\caption{\reviadd{Mean absolute error (MAE) (mean $\pm$ standard deviation) between predicted MRI scans and ground truth scans for three subjects ($n=9$) transitioning from healthy to unhealthy stages, represented by the percentage of total brain volume. Methods requiring longitudinal data or real longitudinal segmentation maps are shown as a reference and marked with an asterisk (*). The best results of the methods requiring a single image are highlighted in \textbf{bold}.}} 
\resizebox{\textwidth}{!}{%
\begin{tabular}{c|ccccccc}
\toprule
\multirow{2}{*}{\textbf{Method}} & \multicolumn{7}{c}{\textbf{Regions} (MAE \% ($\downarrow$))} \\ \cline{2-8} 
    & \textbf{Ventricles} & \textbf{Hippocampi} & \textbf{Amygdalae} & \textbf{Thalami} & \textbf{Caudates} & \textbf{Putamina} & \textbf{CSF} \\ 
\midrule
 Simul@trophy$^*$        & 0.580 (0.534)& 0.021 (0.016)& 0.010 (0.006)& 0.054 (0.027)& 0.015 (0.012)& 0.015 (0.015)& 0.567 (0.580)\\
 Two Visits$^*$          & 0.268 (0.156)& 0.024 (0.026)& 0.014 (0.010)& 0.022 (0.026)& 0.044 (0.032)& 0.017 (0.010)& 0.498 (0.417)\\  \cline{1-8}
 No-PT& 1.033 (0.705)& 0.046 (0.033)& 0.018 (0.011)& 0.086 (0.023)& 0.054 (0.042)& 0.079 (0.038)& \textbf{0.925 (0.845)}\\
 BrLP                    & 0.631 (0.576)& 0.034 (0.033)& 0.013 (0.010)& 0.040 (0.043)& \textbf{0.018 (0.013)}& 0.058 (0.019)& 1.581 (0.764)\\
 InBrainSyn\_HC              & 0.502 (0.545)& 0.026 (0.021)& 0.014 (0.007)& 0.055 (0.021)& 0.038 (0.033)& 0.025 (0.023)& 0.964 (0.886)\\
 InBrainSyn\_AD& \textbf{0.423 (0.514)}& \textbf{0.023 (0.018)}& 0.013 (0.007)& 0.033 (0.016)& 0.021 (0.046)& \textbf{0.017 (0.023)}& 1.109 (0.936)\\
 InBrainSyn\_Inter& 0.446 (0.383)& 0.026 (0.016)& \textbf{0.010 (0.008)}& \textbf{0.028 (0.020)}& 0.049 (0.050)& 0.022 (0.016)& 1.189 (0.917)\\
\bottomrule
\end{tabular}
}
\label{tab:MAE_331}
\end{table*}

\begin{table*}[t]
\renewcommand{\arraystretch}{1.5}
  \begin{center}
    \caption{\reviadd{Evaluation of six similarity metrics (mean $\pm$ standard deviation) for three subjects ($n=9$) transitioning from healthy to unhealthy stages. Methods requiring longitudinal data or real longitudinal segmentation maps are shown as a reference and marked with an asterisk (*). The best results of the methods requiring a single image are highlighted in \textbf{bold}.}}
    \label{tab:6cri_inter}
    \makebox[\textwidth]{%
    \begin{tabular}{c|cccccc}
    \toprule
    \textbf{Method} & \textbf{NFN} ($\downarrow$) & \textbf{MAE} ($\downarrow$)& \textbf{PSNR} ($\uparrow$)& \textbf{SSIM} ($\uparrow$)& \textbf{NCC} ($\uparrow$)& \textbf{DSC} ($\uparrow$)\\
    \midrule
 Simul@trophy$^*$   & 0.059 (0.015) & 0.027 (0.008) & 24.832 (2.278)& 0.898 (0.029) & 0.971 (0.012)& 0.780 (0.080) \\ 
 Two Visits$^*$        & 0.047 (0.013) & 0.022 (0.008) & 26.855 (2.545) & 0.942 (0.024) & 0.986 (0.007) & 0.854 (0.046) \\ \cline{1-7} 
 No-PT& 0.087 (0.006)& 0.039 (0.005)& 21.242 (0.616)& 0.799 (0.010)& 0.925 (0.007)& 0.709 (0.035)\\ 
 BrLP                  & 0.075 (0.013) & 0.034 (0.009) & 22.585 (1.467) & 0.851 (0.020) & 0.952 (0.014) & 0.797 (0.044) \\ 
 InBrainSyn\_HC        & \textbf{0.055 (0.014)} & \textbf{0.025 (0.008)} & \textbf{25.518 (2.247)}& \textbf{0.918 (0.027)} & \textbf{0.976 (0.010)}& \textbf{0.828 (0.061)} \\ 
 InBrainSyn\_AD        & 0.059 (0.012) & 0.026 (0.007) & 24.798 (1.796) & 0.906 (0.024) & 0.972 (0.009) & 0.806 (0.051) \\ 
 InBrainSyn\_Inter     & 0.061 (0.012) & 0.027 (0.007) & 24.439 (1.770) & 0.897 (0.027) & 0.968 (0.011) & 0.814 (0.056) \\ 
 \bottomrule
    \end{tabular}
    }
  \end{center}
\end{table*}

\textbf{Qualitative comparison }: In this section, we randomly select one subject (\textit{OAS30331}) and compare \reviadd{baseline methods} with three settings of our method as shown in Fig. \ref{fig:indiv_compa_331}:
i) $InBrainSyn_{\_}HC$: This setting only uses healthy templates to simulate the subject's evolution over the available time.
ii) $InBrainSyn_{\_}AD$: In this scenario, only AD templates are employed to simulate the subject's evolution over the available time.
iii) $InBrainSyn_{\_}Inter$: This setting considers the transition age to be when the next session results in a non-zero CDR score (i.e., 75.27 for the selected subject). It simulates the evolution using healthy templates before this transition age and switches to AD templates afterwards.
\reviadd{As shown in the highlighted ventricular region (red arrows), the synthetic scans from $InBrainSyn_{\_}Inter$ are closer to the ground truth, as disease-induced changes are accounted for by transitioning from healthy to AD cohort templates after the subject's conversion to AD (CDR$>$0) for this subject}. We note that Simul@tropy \reviadd{and BrLP} tend to generate blurred images, due to performing simulations in a downsampled space.

\reviadd{\textbf{Quantitative  comparison}:}
Table \ref{tab:MAE_331} provides \reviadd{MAE of regional brain volumes across seven regions} \reviadd{for the three subjects ($n=9$) transitioning from healthy to unhealthy stages ($n=3$ for \textit{OAS30331}, $n=2$ for \textit{OAS30869}, and $n=4$ for \textit{OAS30899})}. From the table, we can observe that \reviadd{our proposed method achieves the smallest MAE in most regions and is comparable with the methods that use longitudinal data.}

\reviadd{The results of six similarity metrics are summarized in Table \ref{tab:6cri_inter}. Our proposed InBrainSyn variants demonstrate competitive performance. From Tables \ref{tab:MAE_331} and \ref{tab:6cri_inter}, it is not clear which of the three variants of the proposed InBrainSyn method gives the best results for \textit{both} volume-based and image-based evaluation measures. This might be caused by the limited sample size, the inter-individual heterogeneity of AD, and potential trade-offs between optimal results for each evaluation strategy. However, we note that all variants of InBrainSyn outperform the single-observation baselines.}

\begin{table*}[t]
\renewcommand{\arraystretch}{1.5}
\centering
\caption{\reviiadd{Mean absolute error (MAE) (mean $\pm$ standard deviation) between predicted and ground truth MRI segmentations across brain regions, expressed as a percentage of total brain volume. The table presents the results evaluated on a subset of the ADNI dataset (external set) for both HC and AD cohorts. We selected eight subjects to form each cohort separately. Methods requiring longitudinal data or real longitudinal segmentation maps are shown as a reference and marked with an asterisk (*). The best results of methods requiring a single image are highlighted in \textbf{bold}.}}
\label{tab:adni_external_MAE}
\resizebox{\textwidth}{!}{%
\begin{tabular}{cc|ccccccc}
\toprule
\multirow{2}{*}{\textbf{Cohort}} & \multirow{2}{*}{\textbf{Method}} & \multicolumn{7}{c}{\textbf{Regions} (MAE \% ($\downarrow$))} \\ \cline{3-9}
                        &   & \textbf{Ventricles} & \textbf{Hippocampi} & \textbf{Amygdalae} & \textbf{Thalami} & \textbf{Caudates} & \textbf{Putamina} & \textbf{CSF} \\ 
\midrule
\multirow{5}{*}{\shortstack{\textbf{HC} \\ ($n=25$)}} 
 & Simul@trophy$^*$ &  0.176 (0.158)&  0.008 (0.007)&  0.006 (0.004)&  0.019 (0.011)&  0.010 (0.010)&  0.012 (0.012)&  0.459 (0.356)\\ 
 & Two Visits$^*$ &  0.143 (0.135)&  0.008 (0.005)&  0.004 (0.004)&  0.014 (0.007)&  0.009 (0.007)&  0.010 (0.008)&  0.380 (0.347)\\ \cline{2-9}
 & No-PT &  0.790 (0.741)&  0.039 (0.023)&  0.026 (0.017)&  0.034 (0.036)&  0.035 (0.020)&  0.073 (0.038)&  1.953 (0.525)\\ 
 & BrLP &  0.236 (0.194)&  0.015 (0.010)&  0.018 (0.011)&  0.030 (0.017)&  0.017 (0.012)&  0.019 (0.016)&  0.619 (0.324)\\ 
 & InBrainSyn (Ours) & \textbf{0.148 (0.116)}& \textbf{0.010 (0.007)}& \textbf{0.009 (0.006)}& \textbf{0.011 (0.007)}& \textbf{0.009 (0.006)}& \textbf{0.018 (0.010)}& \textbf{0.265 (0.193)}\\ 
\midrule
\multirow{5}{*}{\shortstack{\textbf{AD} \\ ($n=10$)}}
 & Simul@trophy$^*$ &  0.173 (0.096)&  0.007 (0.005)&  0.009 (0.008)&  0.019 (0.015)&  0.007 (0.005)&  0.013 (0.006)&  0.538 (0.250)\\ 
 & Two Visits$^*$ &  0.080 (0.059)&  0.006 (0.005)&  0.009 (0.007)&  0.018 (0.011)&  0.009 (0.007)&  0.012 (0.008)&  0.326 (0.270)\\ \cline{2-9}
 & No-PT &  1.234 (0.548)&  0.032 (0.026)&  0.028 (0.012)&  0.043 (0.051)&  0.064 (0.047)&  0.053 (0.039)&  1.675 (0.620)\\ 
 & BrLP &  \textbf{0.095 (0.080)}&  0.016 (0.013)&  \textbf{0.014 (0.012)}&  \textbf{0.039 (0.033)}&  \textbf{0.019 (0.014)}&  0.026 (0.021)&  0.627 (0.486)\\ 
 & InBrainSyn (Ours) & 0.102 (0.052) & \textbf{0.013 (0.012)}& 0.018 (0.012) & 0.043 (0.019) & 0.028 (0.015) & \textbf{0.020 (0.009)}& \textbf{0.299 (0.152)}\\ 
\bottomrule
\end{tabular}
}
\end{table*}

\reviiadd{\subsection{Evaluation of Model Generalization}}
\reviiadd{The design of our method allows simulating MRI scans from new datasets with varying MRI protocols and contrasts without requiring retraining of the template generation model. 
To test this potential for generalization, we use a subset of the ADNI dataset \citep{petersen2010alzheimer}. We categorized ADNI subjects into HC and AD cohorts based on PET amyloid and tau biomarkers. Specifically, we defined AD subjects as those who tested positive for both amyloid and tau biomarkers. Tau positivity was determined using the [F-18] Flortaucipir (FTP) tracer. To determine tau positivity, we used the Standardized Uptake Value Ratio (SUVR), a commonly used PET imaging metric that quantifies the relative tracer uptake in a target region compared to a reference region. We focused on the meta-temporal region, a key area for Alzheimer’s-related tau accumulation, and used $\text{META\_TEMPORAL\_SUVR} \geq 1.37$ as the positivity threshold \citep{meyer2020characterization}. Using this stricter selection criterion, we identified eight AD subjects who met these conditions, on top of the criteria established in our previous work \citep{fu2023fast}. To maintain balance, we also randomly selected eight HC subjects, forming two separate cohorts for evaluation. This resulted in 10 MRI scans to simulate for the AD cohort and 25 scans for the HC cohort.}

\reviiadd{For comparison, we evaluated four competing methods as described in Section \ref{sec:eva_individual}, including two reference methods that utilize longitudinal data. Notably, among the baselines, BrLP is partially trained on ADNI, which gives it an inherent advantage in this experiment. While we include BrLP for completeness, it is important to recognize that its results do not reflect true generalization to new datasets. The results are presented in Table~\ref{tab:adni_external_MAE}. As in previous experiments, Two Visits consistently yields good results leveraging longitudinal information. We observe that our single image-only method achieved the best performance on the HC cohort, sometimes even surpassing Two Visits. In turn, in the AD cohort, BrLP performed better in most of the regions, as it was trained partially on ADNI, whereas our method was not.}

\section{Discussion}
In this section, we discuss the results presented in Sect. \ref{sect:results} in terms of cohort-level and individual-level generation, as well as the limitations of the proposed InBrainSyn method.

\subsection{Template Generation}
As shown in Fig. \ref{fig:cohort_templates}, the constructed templates derived from two cohorts are sharp and have well-defined boundaries and robust image contrast. Furthermore, there is a clear progressive enlargement in the lateral ventricles over time, which can also be seen in the residual maps. Across the series of generated deformable templates, residual maps, and the series of difference maps, it is evident that for both the HC cohort and the AD cohort, a discernible enlargement of the ventricles occurs as individuals age. Moreover, according to the residual maps, while both the HC cohort and the AD cohort templates follow a similar deformation pattern with increasing age, the brain morphological changes of the AD cohort are notably more complex and pronounced. When observed from an inter-cohort perspective, it is evident that at each age, the ventricles of the AD cohort exhibit a larger volume than those of the HC cohort. This distinction is particularly pronounced in the early stages of aging, gradually diminishing as individuals grow older\reviiiadd{. This is likely due to the increasing overlap between age-related and AD-related neurodegeneration, where both processes contribute to structural changes over time.} These findings align with established medical literature concerning aging and AD \citep{fox2004imaging,fjell2009one,risacher2010longitudinal}. 

The EFC results are shown in Fig. \ref{fig:efc_for_cohorts}; we can observe that among models, the one employing a $\lambda_{gp}$ value of $10e^{-4}$ exhibits the smallest EFC values for the learned templates, indicating the sharpest templates. Consequently, this model was selected for subsequent experiments.
In Fig. \ref{fig:volumetric_trends}, it is evident that during the aging process, the brain ventricles exhibit an expanding trend, while the hippocampi display a trend of volumetric reduction. For each of the two brain regions of interest, the volumetric trends of the templates of both the HC cohort and the AD cohort are generally aligned with the volumetric trends observed in the real MRI scans in the real database. Furthermore, the curves of templates mostly lie within the variation scope of curves of the scans of the corresponding cohort. Notably, for each region, discernible distinctions emerge between the volumes of the HC cohort and the AD cohort, evident in both real underlying MRI scans and the learned deformable templates. However, such observations are more pronounced at younger ages. Beyond the age of 80 years old, there are more drastic fluctuations in real data, and the curves of the HC and AD MRI scans even overlap severely. This likely arises from the intricacies of neurodegenerative processes. As individuals age, the influence of AD on brain morphology diminishes in comparison to the effects of normal aging \citep{rhodius2017mri,meysami2021quantitative,ouyang2022disentangling,mehta2024early}. This transition in predominant influence affects the learning of templates, as also indicated by the observations from the difference maps. 

\subsection{Individual-Level Image Generation}
\reviiiadd{\textbf{Evaluation on HC and AD Cohorts}}:
Fig. \ref{fig:Individual_comparisons} visually illustrates the comparison results. The first row shows the ground truth longitudinal MRI scans, while the following rows represent benchmark solutions as well as the proposed InBrainSyn in the bottom row. As compared to the No-PT method, the other methods keep personalization of the input subject, while the No-PT method only represents template evolutions. Our approach outperforms Simul@trophy \reviadd{and BrLP} in terms of accuracy in modeling the evolution process and provides superior image resolution with reduced smoothing. \reviadd{While Two Visits generates scans that are most similar to real ones, it requires longitudinal information. In contrast,} the primary objective of InBrainSyn is to address scenarios where only a single visit is available, providing greater flexibility in modeling individual longitudinal changes. \reviadd{From Fig. \ref{fig:Individual_comparisons}, the intensity similarity and aging progression can be assessed on a representative slice for this subject. The figure shows that BrLP and Simul@trophy appear to produce blurrier scans with less noticeable changes over time. To further analyze these temporal changes, Fig. \ref{fig:Individual_trends_comparison} presents volumetric trends for the ventricles and hippocampi in synthetic MRI scans of this subject. The results indicate that BrLP and Simul@trophy exhibit relatively flat volumetric trends for both structures, likely due to the lower resolution of the generated scans, which negatively impacts the performance of the segmentation algorithm. This highlights the importance of resolution in synthetic scans for downstream tasks such as segmentation. The primary reason BrLP and Simul@trophy operate at half resolution (or can only be practically used at half resolution) is the high computational and memory demands required for processing 3D medical images. In contrast, InBrainSyn is unaffected by this limitation as it operates on half-resolution velocity fields while allowing upsampling to full resolution without loss of much information. Typically, downsampling the velocity field introduces significantly less loss compared to directly downsampling images.} 

\reviadd{We also present six similarity metrics in Table \ref{tab:longi_cohort}. The results show that the proposed InBrainSyn consistently generates high-quality MRI scans, outperforming both the No-PT and BrLP. This further validates InBrainSyn in terms of overall synthesis fidelity.}

\reviiiadd{\textbf{Disease Transition Cases}}:
Fig. \ref{fig:indiv_compa_331} visually presents the comparison results. Firstly, all three proposed settings outperform \reviadd{Simul@trophy and BrLP} in terms of superior image resolution with less smoothing, \reviadd{while also surpassing the No-PT in preserving individualization. }Among the three settings, $InBrainSyn_{\_}Inter$ provides the most accurate simulation of the evolution process. \reviadd{This is probably due to its consideration of disease transition during generation. By incorporating inter-cohort template registration, disease-induced changes can be better captured.}
 
\reviadd{Table \ref{tab:MAE_331} and Table \ref{tab:6cri_inter} present the quantitative results for subjects who transitioned from cognitively normal to AD dementia. We observe that InBrainSyn variants achieve the best performance in terms of both similarity and accuracy measures across different regions.  
As expected, the regional MAE results do not indicate an absolute winner among the three variants for AD subjects, given the high heterogeneity of the disease. However, similarity measures clearly identify \textit{InBrainSyn\_HC} as the top-performing variant.  
Upon manually inspecting the three subjects in this experiment, we observed that the subject with the highest number of scans (n=4) exhibited an atypical atrophy pattern—the ventricular region remained relatively stable with minimal changes, while the hippocampal region was more affected. This may have biased the majority of cases toward a pattern more similar to the HC cohort, thereby influencing the similarity measures in Table \ref{tab:6cri_inter}, which evaluate whole scans rather than individual regions as in Table \ref{tab:MAE_331}.}

\subsection{Comparison of Inference Time \reviiadd{and Memory Usage}}

To fairly compare the inference time of the InBrainSyn with other 3D brain MRI simulators, we conducted an experiment using the same machine with Ubuntu 20.04.5 OS and Intel(R) Xeon(R) Bronze 3204 CPU @ 1.90GHz with 64 GB RAM. We \reviiadd{evaluated BrLP, Simul@trophy, Two Visits, and InBrainSyn under CPU-only conditions.} 
The inference steps of the InBrainSyn consist of two template creations ($\sim$26s per creation), two SVF extractions ($\sim$18s per extraction), a parallel transport ($\sim$40s), and the integration ($\sim$2s). In total, the InBrainSyn takes around 2 minutes on the CPU to get a single 3D MRI simulation. If the templates are precomputed, this time is reduced to one minute. The inference time for \reviiadd{Simul@trophy is around 7m30s, though this was measured on undersampled images due to memory constraints. }An issue of Simul@trophy is that it needs to get a prescribed atrophy map. If not available, this map can be estimated from the segmentation differences between two longitudinal scans. \reviiadd{BrLP was tested under two configurations: when running single-threaded, it took approximately 27m, while in its default multi-threaded setting, it required about 23m. Two Visits is the fastest, requiring only 18s, but it also relies on two observations to perform the inference.} 
\reviiadd{Compared to InBrainSyn, BrLP is substantially more time-consuming on CPU, even with multi-threading enabled, due to its diffusion-based generative modeling approach, which relies on iterative sampling ($n=10$ in all our experiments). 
Simul@trophy and Two Visits both benefit from additional longitudinal information, which is not required for InBrainSyn, making it a more flexible solution for cases where only a single baseline scan is available. As a broader reference,} \citet{ravi2022degenerative} reported an inference time of a few minutes for a single 3D MRI simulation using a cluster of GPUs with a total number of 50 NVIDIA GTX TITAN-X. 

\reviiadd{We also measured the peak memory usage for each method. BrLP had the highest memory consumption, reaching 15.7 GB at peak usage on the CPU (around 18 GB at peak usage on the GPU). Simul@trophy used 8.3 GB, while InBrainSyn required 4.5 GB for steps 1 and 3 combined and only 0.2 GB for the parallel transport step. Two Visits had variable memory requirements depending on its configuration: 8.2 GB for the full pipeline and 2.8 GB when disabling stopping point searching. While BrLP and Simul@trophy require significantly more memory, this does not necessarily imply a higher level of computational demand. Memory consumption depends not only on the nature of the task but also on implementation details such as data handling, intermediate storage, and parallelism. InBrainSyn efficiently manages memory by leveraging diffeomorphic transformations rather than storing multiple intermediate states.
In practical deployment, the choice of method depends on the available computational resources and the specific application. BrLP benefits from GPU acceleration, where inference time significantly improves (e.g., ~25s on an NVIDIA RTX A6000 GPU). InBrainSyn offers a balance between efficiency and memory, making it a practical solution for real-world clinical applications with single-scan inference. Two Visits provides the fastest runtime but requires two observations. While memory consumption varies widely across these methods, it is not necessarily a direct indicator of efficiency. Each approach trades off between memory usage, processing time, and required input information, depending on its underlying methodology and design choices.
}
\subsection{Limitations}
The primary limitation encountered is the accurate modeling of AD progression across the cohort. Thus far, we have trained the model on the entire AD cohort to derive a series of deformable templates, which consists of individuals at varying disease stages. These stages encompass subjects with a diagnostic result of AD or DAT as recorded in the OASIS-3 datasets, all grouped under the AD cohort in our study. However, AD is known to follow diverse pathways in terms of morphological changes \citep{poulakis2022multi}, resulting in significant heterogeneity. As illustrated in Fig. \ref{fig:volumetric_trends}, the real volumetric trends within the AD group exhibit substantial noise and fluctuations, especially when compared to the more stable trends observed in the healthy cohort. Furthermore, the scarcity of data in the earlier age groups, such as individuals younger than 65, poses a challenge for the template learning process within the DL model. Consequently, these factors contribute to the relatively flat trends observed in the learned cohort-level AD templates.

The second limitation is associated with \reviiiadd{disease transition cases,}
\reviiiadd{ where we observed less satisfactory cortical generation results when attempting to transport the SVF obtained by registering two templates learned from different cohorts. The primary issue arises because parallel transport assumes that the trajectory being transported should capture only evolutionary changes and not subject-specific differences. However, the template generation model is not designed to produce anatomically similar templates across different cohorts. This issue could be alleviated by jointly learning templates from multiple cohorts when more balanced data across cohorts becomes available.}
\reviiadd{\subsection{Future work}
We identify several promising directions for future research. First, the generated synthetic images can be used for a variety of downstream tasks, such as enhancing data augmentation strategies for deep learning models, improving longitudinal segmentation and registration, and facilitating disease progression modeling in the absence of complete longitudinal data. Second, the proposed approach can be extended to model subtype-specific changes in AD by learning AD subtype templates. An even more direct strategy would involve transporting real individual disease progression patterns from well-established longitudinal datasets. This could be achieved by developing an effective retrieval method that infers subject-specific changes based on a single scan.
}

\section{Conclusions}
In conclusion, this study addresses the challenge of synthesizing individualized high-dimensional brain MRI scans from a single scan.
Current brain simulators face various challenges, including high computational cost, individualization preservation, and explicit anatomical plausibility constraints. To overcome these challenges, we introduced InBrainSyn. In this work, we show that InBrainSyn efficiently synthesizes high-resolution longitudinal MRI scans by combining a deep-learning-based template creation model with a diffeomorphic deformation-based parallel transport algorithm, enabling individual-level synthesis. \reviiadd{We developed our framework using T1w MRI scans from the OASIS-3 dataset and evaluated it on both OASIS-3 and an external subset from the ADNI dataset.} Our evaluations, encompassing quantitative and qualitative analyses, show the effectiveness of InBrainSyn. Our results reveal that InBrainSyn outperforms benchmark models in terms of image quality,
and accuracy, particularly in the context of AD and aging. By design, the use of diffeomorphic registration ensures the anatomical plausibility of the generated images. This study opens new avenues for improving the understanding of neurodegenerative processes and paves the way for more accurate and individualized MRI image synthesis in the field of medical imaging.

\section*{Conflict of Interest}
Dr. D. Ferreira consults for BioArctic and has received honoraria from Esteve. Otherwise, the authors declare no conflict of interest.

\section*{Data Availability Statement}
The data that support the findings of this study are openly available in OASIS at \url{http://doi.org/10.1101/2019.12.13.19014902}, reference number~\citep{lamontagne2019oasis}. The source code \reviiadd{and pretrained networks will be} available at the time of
publication \reviiadd{in our GitHub repository}.

\section*{Funding}  
This study has been partially funded by the Swedish Childhood Cancer Foundation (Barncancerfonden; MT2019-0019, MT2022-0008), by Vinnova through AIDA, project ID: 2108, by the China Scholarship Council (CSC) for PhD studies at KTH Royal Institute of Technology, by Digital Futures, project dBrain, by the Swedish Research Council (Vetenskapsrådet, grants 2022-03389, 2022-00916), MedTechLabs, the Center for Innovative Medicine (CIMED, grants 20200505 and FoUI-988826), the regional agreement on medical training and clinical research of Stockholm Region (ALF Medicine, grants FoUI-962240 and FoUI-987534), the Swedish Brain Foundation (Hjärnfonden FO2023-0261, FO2022-0175, FO2021-0131), the Swedish Alzheimer Foundation (Alzheimerfonden AF-968032, AF-980580), the Swedish Dementia Foundation (Demensfonden), the Gamla Tjänarinnor Foundation, the Gun och Bertil Stohnes Foundation, Funding for Research from Karolinska Institutet, Neurofonden, and Foundation for Geriatric Diseases at Karolinska Institutet. The funders of the study had no role in the study design nor the collection, analysis, and interpretation of data, writing of the report, or decision to submit the manuscript for publication.
 
\section*{Acknowledgements}  
Data were provided in part by OASIS-3: Principal Investigators: T. Benzinger, D. Marcus, J. Morris; NIH P50 AG00561, P30 NS09857781, P01 AG026276, P01 AG003991, R01 AG043434, UL1 TR000448, R01 EB009352. AV-45 doses were provided by Avid Radiopharmaceuticals, a wholly-owned subsidiary of Eli Lilly. 


\bibliographystyle{model2-names.bst}\biboptions{authoryear}
\bibliography{references}



\end{document}